\documentclass[10pt,twocolumn,letterpaper]{article}

\usepackage{iccv}
\usepackage{times}
\usepackage{graphicx}
\usepackage{amsmath}
\usepackage{amssymb}
\usepackage{booktabs} % for much better looking tables
\usepackage{array} % for better arrays (eg matrices) in maths
\usepackage{paralist} % very flexible & customisable lists (eg. enumerate/itemize, etc.)
\usepackage{verbatim} % adds environment for commenting out blocks of text & for better verbatim

\usepackage{stmaryrd,bm,abraces}
\usepackage{color}
\usepackage{float}
\usepackage{array}
\usepackage{makecell}
\usepackage{multirow}
\usepackage{verbatim}
\usepackage{soul}
\usepackage{float}
\usepackage{enumitem}
\usepackage{arydshln}
\usepackage{mathtools}
\usepackage{threeparttable}
\usepackage[lined,ruled,commentsnumbered]{algorithm2e}
\usepackage{lipsum}

\usepackage{xcolor}
\usepackage{caption}
\usepackage{afterpage}

\usepackage{environ}
\NewEnviron{lequation}{
\begin{equation}
\scalebox{1.1}{$\BODY$}
\end{equation}
}

\newcolumntype{v}{>{\centering\arraybackslash}m{1.07cm}}
\newcolumntype{y}{>{\centering\arraybackslash}m{0.05cm}}
\newcolumntype{L}{>{\centering\arraybackslash}m{2.3cm}}
\newcolumntype{X}{>{\centering\arraybackslash}m{3.2cm}}
\newcolumntype{Y}{>{\centering\arraybackslash}m{2.7cm}}
\newcolumntype{S}{>{\centering\arraybackslash}m{3.0cm}}
\newcolumntype{F}{>{\centering\arraybackslash}m{4.0cm}}
\newcolumntype{K}{>{\centering\arraybackslash}m{3.7cm}}
\newcolumntype{N}{>{\centering\arraybackslash}m{0.6cm}}
\newcolumntype{Z}{>{\centering\arraybackslash}m{0.1cm}}
\newcolumntype{w}{>{\centering\arraybackslash}m{1.9cm}}
\newcolumntype{p}{>{\centering\arraybackslash}m{1.2cm}}

\newcommand{\mihash}{$\mathsf{MIHash}$}

\newcommand{\vp}{\varoplus}
\newcommand{\vm}{\varominus}
\newcommand{\xhat}{{\hat{\mathbf{x}}}}
\newcommand{\x}{{\mathbf{x}}}
\newcommand{\C}{\mathcal{C}}
\newcommand{\D}{\mathcal{D}}
\usepackage[breaklinks=true,bookmarks=false]{hyperref}

\iccvfinalcopy % *** Uncomment this line for the final submission

\def\iccvPaperID{158} % *** Enter the ICCV Paper ID here

\ificcvfinal\fi
\begin{document}

% ------------------------- Title
% ------------------------- Title
% ------------------------- Title
\title{$\mathbf{\mathsf{MIHash}}$: Online Hashing with Mutual Information}

\author{
Fatih Cakir\thanks{First two authors contributed equally.} ~~~~ Kun He\footnotemark[1] ~~~~ Sarah Adel Bargal ~~~ Stan Sclaroff\\
Department of Computer Science\\
Boston University, Boston, MA\\
{\tt\small \{fcakir,hekun,sbargal,sclaroff\}@cs.bu.edu}
}
\maketitle
\thispagestyle{empty}

% ------------------------- Abstract
% ------------------------- Abstract
% ------------------------- Abstract
\begin{abstract}
Learning-based hashing methods are widely used for nearest neighbor retrieval, and recently, online hashing methods have demonstrated  good performance-complexity trade-offs by learning hash functions from streaming data.
In this paper, 
we first address a key challenge for online hashing: the binary codes for indexed data must be recomputed to keep pace with updates to the hash functions.
We propose an efficient quality measure for hash functions, based on an information-theoretic quantity, \emph{mutual information}, and use it successfully as a criterion to eliminate unnecessary hash table updates.
Next, we also show how to optimize the mutual information objective using stochastic gradient descent. 
We thus develop a novel hashing method, \mihash, that can be used in both online and batch settings.
Experiments on image retrieval benchmarks (including a 2.5M image dataset) confirm the effectiveness of our formulation, both in reducing hash table recomputations and in learning high-quality hash functions.
\end{abstract}

% ------------------------- Introduction
% ------------------------- Introduction
% ------------------------- Introduction

\vspace*{-0.67em}
\section{Introduction}
\vspace*{-0.33em}
\label{sec:intro}
Hashing is a widely used approach for practical nearest neighbor search in many computer vision applications. 
It has been observed that {adaptive} hashing methods that learn to hash from data generally outperform data-independent hashing methods such as Locality Sensitive Hashing  \cite{LSH1}.
In this paper, we focus on a relatively new family of adaptive hashing methods, namely \emph{online} adaptive hashing methods \cite{cakir2015adaptive,OECC,OKH,osketchhash}. These techniques employ online learning in the presence of streaming data, and are appealing due to their low computational complexity and their ability to adapt to changes in the data distribution. 
 
Despite recent progress, a key challenge has not been addressed in online hashing, which motivates this work: the computed binary representations, or the ``hash table", 
may become outdated after a change in the hash mapping. To reflect the updates in the hash mapping, the hash table may need to be recomputed frequently, causing inefficiencies in the system such as successive disk I/O, especially when dealing with large datasets. 
We thus identify an important question for online adaptive hashing systems: 
\textit{when to update the hash table?} 
Previous online hashing solutions do not address this question, as they usually update both the hash mapping and hash table concurrently. 

% ------------------------- fig
% ------------------------- fig
\begin{figure}[t]
\centering
\includegraphics[width=.9\linewidth]{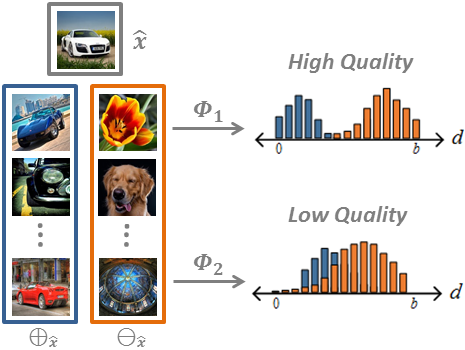}
\caption{We study online hashing  for efficient nearest neighbor retrieval.
Given a hash mapping $\Phi$, an image $\xhat$, along with its neighbors in $\vp_\xhat$ and non-neighbors in $\vm_\xhat$, are mapped to binary codes, yielding two distributions of Hamming distances. In this example, $\Phi_1$ has higher quality than $\Phi_2$ since it induces more separable distributions.
The information-theoretic quantity Mutual Information can be used to capture the separability, which gives a good quality indicator and learning objective for online hashing. 
}
\vspace{-1em}
\label{fig:mi}
\end{figure} 

We make the observation that achieving high quality nearest neighbor search is an ultimate goal in hashing systems, and therefore any effort to limit computational complexity should preserve, if not improve, that quality.
Therefore, another important question is: \textit{how to quantify quality?}
Here, we briefly describe our answer to this question, but first introduce some necessary notation.
We would like to learn a {hash mapping} $\Phi$ from feature space $\mathcal{X}$ to the $b$-dimensional Hamming space $\mathcal{H}^b$, whose outputs are $b$-bit binary codes.
The goal of hashing is to preserve a {neighborhood structure} in $\mathcal{X}$ after the mapping to $\mathcal{H}^b$.  Given $\xhat\in\mathcal{X}$, the neighborhood structure is usually given in terms of a set of its neighbors $\vp_\xhat$, and a set of non-neighbors $\vm_\xhat$. We discuss how to derive the neighborhood structure in Sec.~\ref{sec:formulation}.

As shown in Fig.~\ref{fig:mi}, the distributions of the Hamming distances from $\xhat$ to its neighbors and non-neighbors are histograms over $\{0,1,\ldots,b\}$.
Ideally, if there is no overlap between these two distributions, we can recover $\vp_\xhat$ and $\vm_\xhat$ by simply thresholding the Hamming distance. A nonzero overlap results in ambiguity, as observing the Hamming distance is no longer sufficient to determine neighbor relationships.
Our discovery is that this overlap can be quantified using an information-theoretic quantity, \emph{mutual information}, between two random variables induced by $\Phi$.
We then use mutual information to define a novel measure to quantify quality for hash functions in general.

With a quality measure defined, we answer the motivating question of when to update the hash table. 
We  propose a simple solution by restricting updates to times when there is an estimated improvement in hashing quality, based on an efficient estimation method in the presence of streaming data.
Notably, since mutual information is a good general-purpose quality measure for hashing, this results in a general {plug-in module} for online hashing that does not require knowledge of the learning method. 

Inspired by this strong result, we next ask, \textit{can we optimize mutual information as an objective to learn hash functions?} 
We propose a novel hashing method, \mihash, by deriving gradient descent rules on the mutual information objective, which can be applied in online stochastic optimization, as well as on deep architectures.
The mutual information objective is free of tuning parameters, unlike others
that may require thresholds, margins, \etc.

We conduct experiments on three image retrieval benchmarks, including the Places205 dataset \cite{zhou2014learning} with 2.5M images.
For four recent online hashing methods, our mutual information based update criterion consistently leads to over an order of magnitude reduction in hash table recomputations, while maintaining retrieval accuracy. 
Moreover, our novel \mihash~method achieves state-of-the-art retrieval results,
in both online and batch learning settings.

% ------------------------- Related Work
% ------------------------- Related Work
% ------------------------- Related Work
\vspace*{-0.45em}
\section{Related Work}
\vspace*{-0.25em}
In this section, we mainly review hashing methods that adaptively update the hash mapping with incoming data, given that our focus is on online adaptive hashing. For a more general survey on hashing, please refer to \cite{hash_survey}.

Huang \etal \cite{OKH} propose Online Kernel Hashing, where a stochastic environment is considered with pairs of points arriving sequentially. At each step, a number of hash functions are selected based on a Hamming loss measure and parameters are updated via stochastic gradient descent (SGD).  

Cakir and Sclaroff \cite{cakir2015adaptive} 
argue that, in a stochastic setting, it is difficult to determine which hash functions to update as it is the collective effort of all the hash functions that yields a good hash mapping. Hamming loss is considered to infer the hash functions to be updated at each step and a squared error loss is minimized via SGD. 

In \cite{OECC}, binary Error Correcting Output Codes (ECOCs) are employed in learning the hash functions.  This work follows a more general two-step hashing framework \cite{FastHash}, where the set of ECOCs are generated beforehand and are assigned to labeled data as they appear, allowing the label space to grow with incoming data.
Then, hash functions are learned to fit the binary ECOCs using Online Boosting.

Inspired by the idea of ``data sketching", Leng \etal introduce Online Sketching Hashing \cite{osketchhash} where a small fixed-size sketch of the incoming data is maintained in an online fashion. 
The sketch retains the Frobenius norm of the full data matrix, which offers space savings, and allows to apply certain batch-based hashing methods.
A PCA-based batch learning method is applied on the sketch to obtain hash functions.

None of the above online hashing methods offer a solution to decide whether or not the hash table shall be updated given a new hash mapping. However,  such a solution is crucial in practice, as limiting the frequency of updates will alleviate the computational burden of keeping the hash table up-to-date.
Although  \cite{OECC} and \cite{OKH} include strategies to select individual hash functions to recompute, they still require computing on all indexed data instances.

Recently, some methods employ deep neural networks to learn hash mappings, \eg \cite{Li_IJCAI2016,Lin_CVPR2016,Wang_ACCV2016,Ziming_VeryDeep_CVPR2016} and others. 
These methods use minibatch-based stochastic optimization, however, they usually require multiple passes over a given dataset to learn the hash mapping, and the hash table is only computed  when the hash mapping has been learned. 
Therefore, current deep learning based hashing methods are essentially batch learning methods, which differ from the online hashing methods that we consider, \ie methods that process streaming data to learn and update the hash mappings on-the-fly while continuously updating the hash table.
Nevertheless, when evaluating our mutual information based hashing objective, we compare against state-of-the-art batch hashing formulations as well, by contrasting different objective functions on the same model architecture.

Lastly, Ustinova \etal \cite{ustinova2016learning} recently proposed a method to derive differentiation rules for objective functions that require histogram binning, and apply it in learning deep embeddings. When optimizing our mutual information objective, we utilize their differentiable histogram binning technique for deriving gradient-based optimization rules.
Note that both our problem setup and objective function are quite different from \cite{ustinova2016learning}.

% ------------------------- Formulation
% ------------------------- Formulation
% ------------------------- Formulation
\section{Online Hashing with Mutual Information}
\label{sec:formulation}

% ------------------------- Problem Statement
As mentioned in Sec.~\ref{sec:intro}, the goal of hashing is to learn a hash mapping $\Phi: \mathcal{X} \to \mathcal{H}^b$ such that a desired neighborhood structure is preserved. 
We consider an online learning setup where $\Phi$ is continuously updated from input streaming data, and at time $t$, the current mapping $\Phi_t$ is learned from $\{\mathbf{x}_{1},\ldots,\mathbf{x}_{t}\}$.
We follow the standard setup of  learning $\Phi$ from pairs of instances with similar/dissimilar labels \cite{BRE, OKH, cakir2015adaptive, Li_IJCAI2016}.
These labels, along with the neighborhood structure, can be derived from a metric, \eg two instances are labeled similar (\ie neighbors of each other) if their Euclidean distance in $\mathcal{X}$ is below a threshold. Such a setting is often called unsupervised hashing.
On the other hand, in supervised hashing with labeled data, pair labels are derived from individual class labels: instances are similar if they are from the same class, and dissimilar otherwise.

Below, we first derive the mutual information quality measure and discuss its use in determining when to update the hash table in Sec.~\ref{sec:trigger_update}. We then describe a gradient-based approach for optimizing the same quality measure, as an objective for learning hash mappings, in Sec.~\ref{sec:mi_objective}. Finally, we discuss the benefits of using mutual information in Sec.~\ref{sec:mi_benefits}.

% ------------------------- Update Criterion
\subsection{MI as Update Criterion}
\label{sec:trigger_update}
We revisit our motivating question: \emph{When to update the hash table in online hashing?}
During the online learning of $\Phi_t$, we assume a retrieval set $\mathcal{S}\subseteq\mathcal{X}$, which may include the streaming data after they are received.
We define the {hash table} as the set of hashed binary codes: $\mathcal{T}(\mathcal{S},\Phi)=\{\Phi(\mathbf{x})|\mathbf{x}\in\mathcal{S}\}$.
Given the adaptive nature of online hashing, $\mathcal{T}$ may need to be recomputed often to keep pace with $\Phi_t$; however,  this is undesirable if $\mathcal{S}$ is large or the change in $\Phi_t$'s quality does not justify the cost of an update. 

We propose to view the learning of $\Phi_t$ and computation of $\mathcal{T}$ as separate events, which may happen at different rates. 
To this end, we introduce the notion of a \emph{snapshot}, denoted $\Phi^s$, which is occasionally taken of $\Phi_t$ and used to recompute $\mathcal{T}$. Importantly, this happens only when the nearest neighbor retrieval quality of $\Phi_t$ has improved, and we now define the quality measure.

Given hash mapping $\Phi:\mathcal{X}\to\{-1,+1\}^b$, $\Phi$ induces Hamming distance $d_\Phi:\mathcal{X}\times\mathcal{X}\to\{0,1,\ldots,b\}$ as
\begin{align}
d_\Phi(\mathbf{x}, \xhat) = \frac{1}{2}\left(b - \Phi(\mathbf{x})^\top\Phi(\xhat)\right).
\label{eq:d_phi}
\end{align}
Consider some instance $\mathbf{\hat{x}}\in\mathcal{X}$, and the
sets containing neighbors and non-neighbors, $\varoplus_{\hat{\mathbf{x}}}$ and $\varominus_{\hat{\mathbf{x}}}$.
$\Phi$ induces two conditional distributions, $P(d_\Phi(\mathbf{x},\mathbf{\hat{x}})|\mathbf{x}\in \varoplus_{\hat{\mathbf{x}}})$ and $P(d_\Phi(\mathbf{x},\mathbf{\hat{x}})|\mathbf{x}\in \varominus_{\hat{\mathbf{x}}})$ as seen in Fig.~\ref{fig:mi},
and it is desirable to have low overlap between them.
To formulate the idea, for $\Phi$ and $\mathbf{\hat{x}}$, define random variable $\mathcal{D}_{\mathbf{\hat{x}},\Phi}\!\!:\!\mathcal{X}\!\to\!\{0,1,\ldots,b\}, \x\mapsto d_\Phi(\mathbf{x}, \mathbf{\hat{x}})$, and let $\mathcal{C}_{\hat{\mathbf{x}}}\!\!:\!\mathcal{X}\!\to\!\{0,1\}$  be the membership indicator for $\vp_\xhat$.
The two conditional distributions can now be expressed as 
$P(\D_{\xhat,\Phi}|\C_\xhat=1)$ and $P(\D_{\xhat,\Phi}|\C_\xhat=0)$, and we can write the \emph{mutual information} between $\mathcal{D}_{\mathbf{\hat{x}},\Phi}$ and $\mathcal{C}_{\mathbf{\hat{x}}}$ as
\begin{align}
\mathcal{I}(\mathcal{D}_{\mathbf{\hat{x}},\Phi}; \mathcal{C}_{\mathbf{\hat{x}}}) 
& = H(\mathcal{C}_{\mathbf{\hat{x}}}) - H(\mathcal{C}_{\mathbf{\hat{x}}} | \mathcal{D}_{\mathbf{\hat{x}},\Phi})\\
& = H(\D_{\xhat,\Phi})-H(\D_{\xhat,\Phi}|\C_{\xhat})
\label{eq:mi}
\end{align}
where $H$ denotes (conditional) entropy.
In the following, for brevity we will drop subscripts $\Phi$ and $\hat{\mathbf{x}}$, and denote the two conditional distributions and the marginal $P(\D_{\xhat,\Phi})$ as $p_\D^+$, $p_\D^-$, and $p_\D$, respectively.

By definition, $\mathcal{I}(\D;\C)$  measures the decrease in uncertainty in the neighborhood information $\mathcal{C}$ when observing the Hamming distances $\mathcal{D}$.
We claim that $\mathcal{I}(\D; \C)$ also captures how well $\Phi$ preserves the neighborhood structure of $\hat{\mathbf{x}}$.
If $\mathcal{I}(\D;\C)$ attains a high value, which means $\mathcal{C}$ can be determined with low uncertainty by observing $\mathcal{D}$, then $\Phi$ must have achieved good separation (\ie low overlap) between $p_{\mathcal{D}}^+$ and $p_{\mathcal{D}}^-$.
$\mathcal{I}$ is maximized when there is no overlap, and minimized when $p_{\mathcal{D}}^+$ and $p_{\mathcal{D}}^-$ are exactly identical.
Recall, however, that $\mathcal{I}$ is defined with respect to a single instance $\xhat$; therefore, for a general quality measure, we integrate $\mathcal{I}$ over  the feature space: 
\begin{equation}
Q(\Phi) = \int_{\mathcal{X}} \mathcal{I}({\mathcal{D}_{\mathbf{\hat{x}},\Phi}};C_{\mathbf{\hat{x}}})p(\mathbf{\hat{x}})d\mathbf{\hat{x}}.
\label{mi_batch_criteria}
\end{equation} 
$Q(\Phi)$ captures the expected amount of separation between $p_{\mathcal{D}}^+$ and $p_{\mathcal{D}}^-$ achieved by $\Phi$, over all instances in $\mathcal{X}$.

In the online setting, given the current hash mapping $\Phi_t$ and previous snapshot $\Phi^s$,  it is then straightforward to pose the update criterion as
\begin{align}
Q(\Phi_t)-Q(\Phi^s) > \theta,
\label{mi_step_criteria}
\end{align}
\noindent where $\theta$ is a threshold; a straightforward choice is $\theta=0$.
However, Eq.~\ref{mi_batch_criteria} is generally difficult to evaluate due to the intractable integral; in practice, we resort to Monte-Carlo approximations to this integral, as we describe next.
 
% -------------------- reservoir
% -------------------- reservoir
\vspace{-1em}
\subsubsection*{Monte-Carlo Approximation by Reservoir Sampling} 
\vspace{-.5em}
We give a Monte-Carlo approximation of Eq.~\ref{mi_batch_criteria}. Since we work with streaming data, we employ the Reservoir Sampling algorithm \cite{vitter85}, which enables sampling from a stream or sets of large/unknown cardinality. 
With reservoir sampling, we obtain a \emph{reservoir set} $\mathcal{R}\triangleq\{\mathbf{x}_1^r,\ldots,\mathbf{x}^r_K\}$ from the stream, which can be regarded as a finite sample from $p(\mathbf{x})$. 
We estimate the value of $Q$ on $\mathcal{R}$ as:
\begin{align} 
Q_{\mathcal{R}}(\Phi)  =\frac{1}{|\mathcal{R}|} \sum_{\mathbf{x^r} \in \mathcal{R}} \mathcal{I}_{\mathcal{R}}({\mathcal{D}_{\mathbf{x^r},\Phi}};\mathcal{C}_{\mathbf{x^r}}).
\label{eq:Q_R}
\end{align}
We use subscript $\mathcal{R}$ to indicate that when computing the mutual information $\mathcal{I}$, the $p_\D^+$ and $p_\D^-$ for a reservoir instance $\mathbf{x^r}$ are estimated from $\mathcal{R}$.
This can be done in $\mathcal{O}(|\mathcal{R}|)$ time for each $\mathbf{x^r}$, as the discrete distributions can be estimated via histogram binning.

\begin{figure}[t]
\centering
\includegraphics[width=0.94\linewidth]{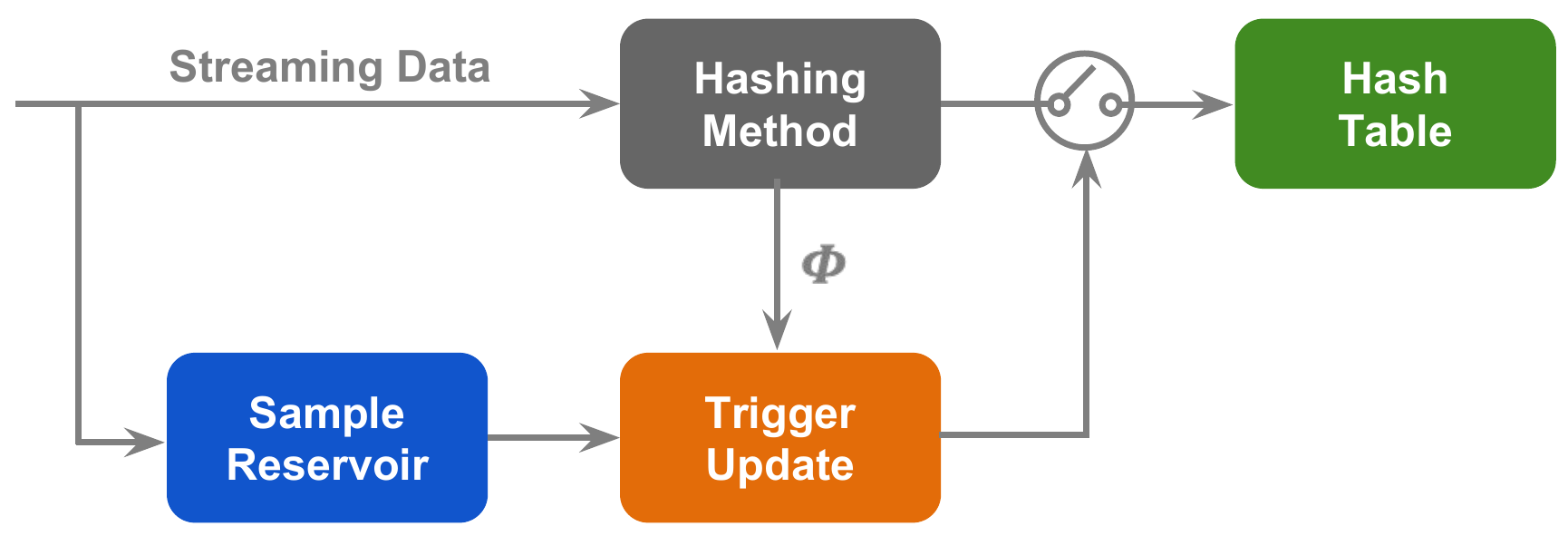}
\vspace{-0.2em}
\caption{We present the general plug-in module for online hashing methods: Trigger Update ({\tt TU}). We sample a reservoir $\mathcal{R}$ from the input stream, and estimate the mutual information criterion $Q_{\mathcal{R}}$. Based on its value, {\tt TU} decides whether a hash table update should be executed.}
\label{fig:TU_plugin}
\end{figure}

Fig.~\ref{fig:TU_plugin} summarizes our approach. We use the reservoir set to estimate the quality $Q_\mathcal{R}$, and ``trigger" an update to the hash table only when $Q_\mathcal{R}$ improves over a threshold. 
Notably, our  approach provides a general \emph{plug-in module} for online hashing techniques, in that it only needs access to streaming data and the hash mapping itself, independent of the hashing method's inner workings.

% ------------------------- fig

% ------------------------- Objective Function
% ------------------------- Objective Function
% ------------------------- Objective Function
\subsection{MI as Learning Objective}
\label{sec:mi_objective}
Having shown that mutual information is a suitable measure of neighborhood quality, we consider its use as a learning objective for {hashing}. 
Following the notation in Sec.~\ref{sec:trigger_update}, we define a loss $\mathcal{L}$ with respect to $\xhat\in\mathcal{X}$ and $\Phi$ as
\begin{align}
\mathcal{L}(\xhat,\Phi)=-\mathcal{I}(\D_{\xhat,\Phi}; \C_\xhat).
\label{eq:mi_obj}
\end{align}
We model $\Phi$ as a collection of parameterized {hash functions}, each responsible for generating a single bit: $\Phi(\mathbf{x}) = [\phi_1 (\mathbf{x};W), ..., \phi_b (\mathbf{x};W)]$, where $\phi_i:\mathcal{X}\rightarrow\{-1,+1\},\forall i$, and $W$ represents the model parameters.  
For example, linear hash functions can be written as $\phi_i(\x) = \text{sgn} (w_i^\top \mathbf{x})$, and for deep neural networks the bits are generated by thresholding the activations of the output layer.

Inspired by the online nature of the problem and recent advances in stochastic optimization, we derive gradient descent rules for $\mathcal{L}$.
The entropy-based mutual information is differentiable with respect to the entries of $p_\D$, $p_\D^+$ and $p_\D^-$, and, as mentioned before, these discrete distributions can be estimated via histogram binning. However, it is not clear how to differentiate histogram binning to generate gradients for model parameters. We describe a differentiable histogram binning technique next.

% ----------------------- histogram binning
% ----------------------- histogram binning
\vspace{-1em}
\subsubsection*{Differentiable Histogram Binning}
\vspace{-.5em}
We borrow ideas from \cite{ustinova2016learning} and  estimate $p_\D^+$, $p_\D^-$ and $p_\D$ using a differentiable histogram binning technique.
For $b$-bit Hamming distances, we use $(K+1)$-bin normalized histograms with bin centers {$v_0 = 0, ..., v_K = b$} and uniform bin width $\Delta = b/K$, where $K=b$ by default.
Consider, for example, the $k$-th entry in $p^+_\D$, denoted as $p_{\mathcal{D},k}^+$. It can be estimated as
\begin{equation}
p_{\mathcal{D},k}^+ = \frac{1}{|\vp|} \sum_{\x\in\vp} \delta_{\x,k},
\label{eq:delta}
\end{equation}
where $\delta_{\x,k}$ records the contribution of $\x$ to bin $k$. 
It is obtained by interpolating $d_\Phi(\x,\xhat)$ using a triangular kernel:
\begin{equation}
  \delta_{\x,k} =
  \begin{cases}
  (d_\Phi(\mathbf{x}, \xhat) - v_{k-1})/\Delta, & d_\Phi(\mathbf{x}, \xhat) \in [v_{k-1}, v_k], \\
  (v_{k+1} - d_\Phi(\mathbf{x}, \xhat))/\Delta, & d_\Phi(\mathbf{x}, \xhat) \in [v_{k}, v_{k+1}], \\
  0, & \text{otherwise.}
  \end{cases}
\end{equation}
This binning process admits subgradients:
\begin{equation}
  \frac{\partial \delta_{\x, k}}{\partial d_\Phi(\mathbf{x}, \xhat)} =
  \begin{cases}
  1/\Delta, & d_\Phi(\mathbf{x}, \xhat) \in [v_{k-1}, v_k], \\
  -1/\Delta, & d_\Phi(\mathbf{x}, \xhat) \in [v_{k}, v_{k+1}], \\
  0, & \text{otherwise.}
  \end{cases}
  \label{eq:delta-dh}
\end{equation}

%---------------------- fig
%---------------------- fig
\begin{figure*}[t!]
\centering
~\includegraphics[width=0.325\linewidth]{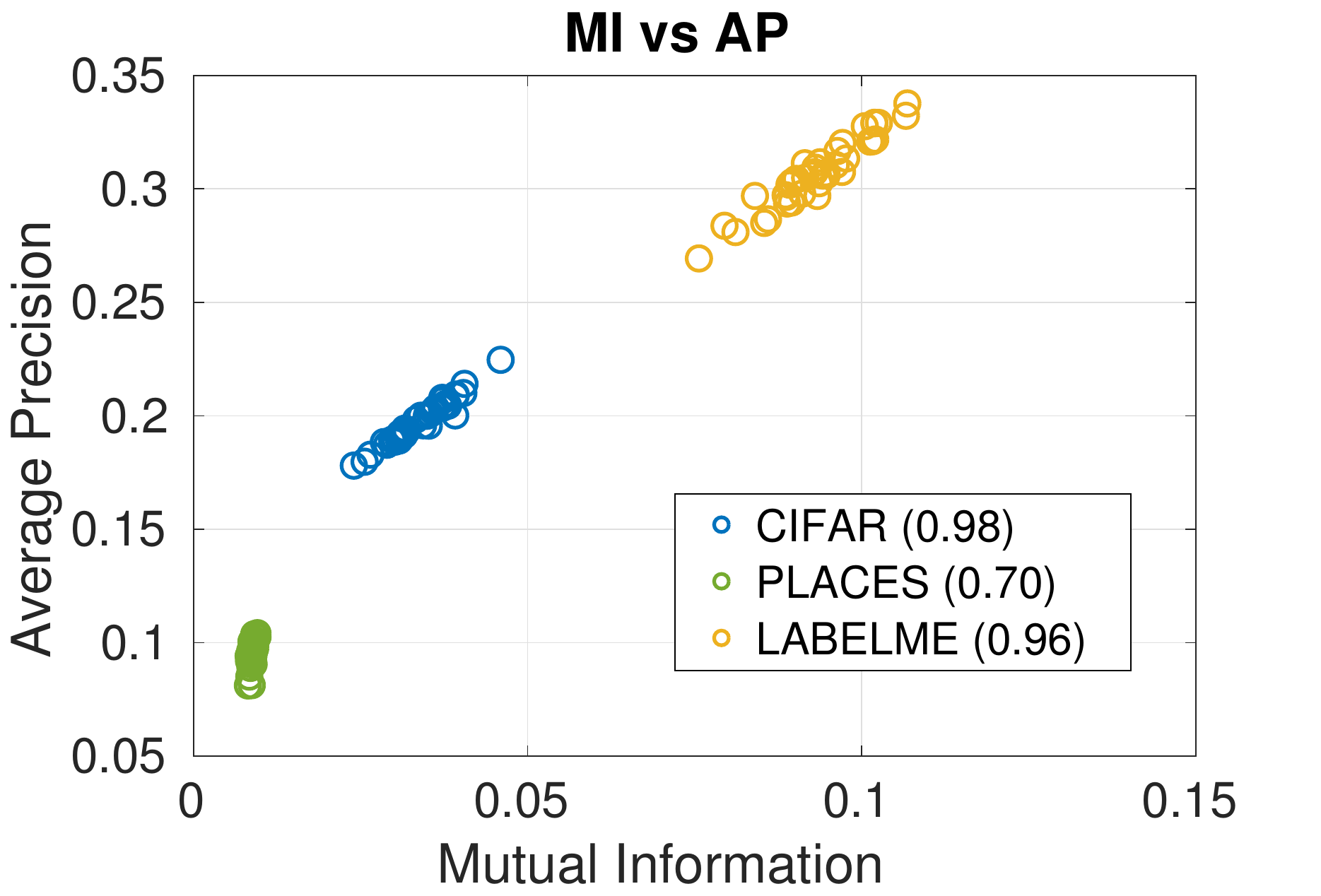}
\includegraphics[width=0.325\linewidth]{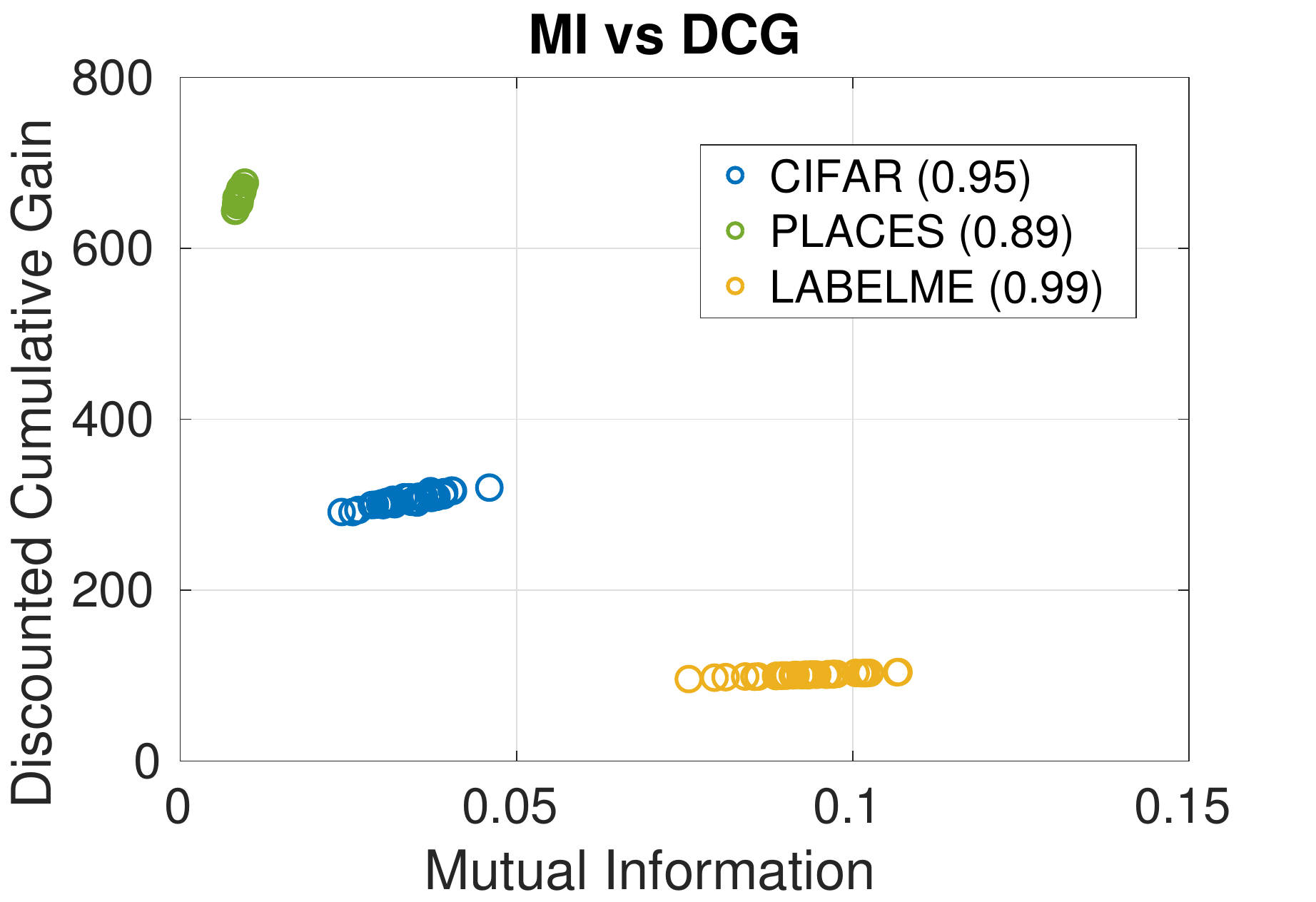}
\includegraphics[width=0.325\linewidth]{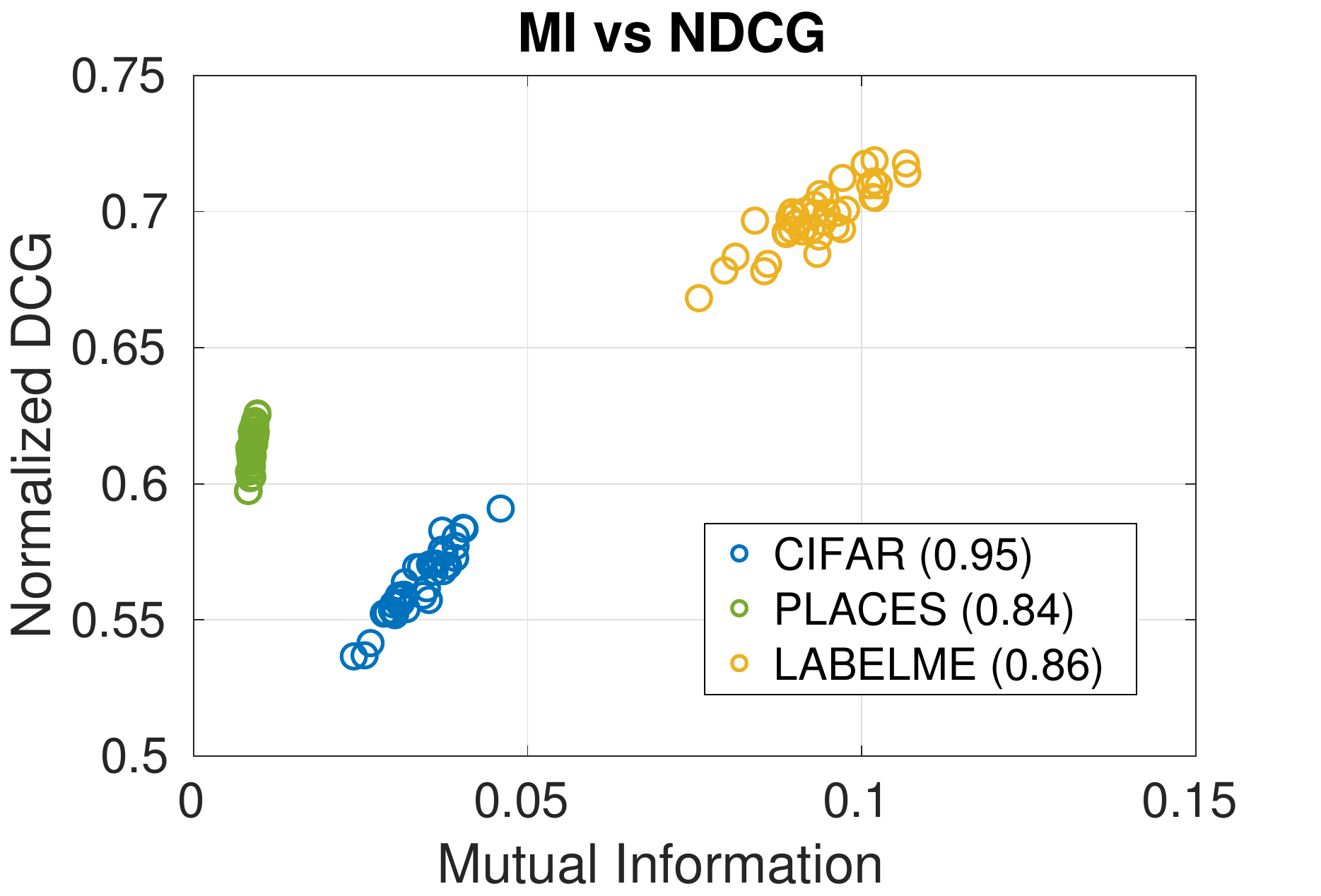}
\vspace{-.25em}
\caption{We show Pearson correlation coefficients between mutual information  (MI) and AP, DCG, and NDCG, evaluated on the CIFAR-10, LabelMe, and Places205 datasets.  
We sample $100$ instances to form the query set, and use the rest to populate the hash table. 
The hash mapping parameters are randomly sampled from a Gaussian, similar to LSH \cite{LSH1}.  Each experiment is conducted $50$ times.
There exist strong correlations between MI and the standard metrics.
}
\label{fig:correlation}
\end{figure*}

% ----------------------- SGD MI
% ----------------------- SGD MI
\vspace{-1em}
\subsubsection*{Gradients of MI}
\vspace{-.5em}
We now derive the gradient of $\mathcal{I}$ with respect to the output of the hash mapping, $\Phi(\xhat)$.
Using standard chain rule, we can first write
\begin{align}
\frac{\partial \mathcal{I}}{\partial \Phi(\xhat)}=\sum_{k=0}^K\left[\frac{\partial \mathcal{I}}{\partial p^+_{\D,k}}\frac{\partial p^+_{\D,k}}{\partial \Phi(\xhat)}
+ \frac{\partial \mathcal{I}}{\partial p^-_{\D,k}}\frac{\partial p^-_{\D,k}}{\partial \Phi(\xhat)}\right].\label{eq:diff_mi}
\end{align}

We focus on terms involving $p_{\D,k}^+$, and omit derivations for $p_{\D,k}^-$ due to symmetry. For $k=0,\ldots,K$, we have
\begin{align}
\frac{\partial \mathcal{I}}{\partial p^+_{\D,k}} & =
-\frac{\partial H(\D|\C)}{\partial p^+_{\D,k}}+\frac{\partial H(\D)}{\partial p^+_{\D,k}}\\
& = p^+(\log p^+_{\D,k}+1)-(\log p_{\D,k}+1)\frac{\partial p_{\D,k}}{\partial p^+_{\D,k}}\\
& = p^+(\log p^+_{\D,k}-\log p_{\D,k}),
\end{align}
where we used the fact that $p_{\D,k}=p^+p^+_{\D,k}+p^-p^-_{\D,k}$, with $p^+$ and $p^-$ being shorthands for the priors $P(\C=1)$ and $P(\C=0)$.
We next tackle the term $\partial p^+_{\D,k}/\partial\Phi(\xhat)$ in Eq.~\ref{eq:diff_mi}. From the definition of $p_{\D,k}^+$ in Eq.\ref{eq:delta}, we have
\begin{align}
\frac{\partial p_{\mathcal{D},k}^+}{\partial \Phi(\xhat)}
&= \frac{1}{|\vp|} \sum_{\mathbf{x} \in \vp} \frac{\partial \delta_{\mathbf{x},k}}{\partial \Phi(\xhat)}\\
&= \frac{1}{|\vp|} \sum_{\mathbf{x} \in \vp} \frac{\partial \delta_{\mathbf{x},k}}{\partial d_\Phi(\x,\xhat)}\frac{\partial d_\Phi(\x,\xhat)}{\partial \Phi(\xhat)}\\
&= \frac{1}{|\vp|} \sum_{\mathbf{x} \in \vp} \frac{\partial \delta_{\mathbf{x},k}}{\partial d_\Phi(\x,\xhat)}\frac{-\Phi(\x)}{2}.
\label{eq:condprob-phi}
\end{align}
Note that $\partial\delta_{\x,k}/\partial d_\Phi(\x,\xhat)$ is already given in Eq. \ref{eq:delta-dh}. For the last step, we used the definition of $d_\Phi$ in Eq. \ref{eq:d_phi}.

Lastly, to back-propagate gradients to $\Phi$'s {inputs} and ultimately model parameters, we approximate the discontinuous sign function with  sigmoid, which is a standard technique in hashing, \eg \cite{cakir2015adaptive,Li_IJCAI2016,SHK}.

% ------------------------- Correlation
% ------------------------- Correlation
% ------------------------- Correlation
\subsection{Benefits of MI} 
\label{sec:mi_benefits}
%\vspace*{-0.5em}
For monitoring the performance of hashing algorithms, it appears that one could directly use 
standard ranking metrics, such as Average Precision (AP), Discounted Cumulative Gain (DCG), and Normalized DCG (NDCG) \cite{Manning_irbook2008}.
Here, we discuss the benefits of instead using mutual information.
First, we note that there exist strong correlations between mutual information and standard ranking metrics. Fig.~\ref{fig:correlation} demonstrates the Pearson correlation coefficients between MI and AP, DCG, and NDCG, on three benchmarks. Although a theoretical analysis is beyond the scope of this work, empirically we find that MI serves as an efficient and general-purpose ranking surrogate.

We also point out the lower computational complexity of mutual information.
Let $n$ be the reservoir set size. Computing Eq.~\ref{eq:Q_R} involves estimating discrete distributions via histogram binning, and takes $\mathcal{O}(n)$ time for each reservoir item, since $\D$ only takes discrete values from $\{0,1,\ldots,b\}$,
In contrast, ranking measures such as AP and NDCG have $\mathcal{O}(n\log n)$ complexity due to sorting, which render them disadvantageous.

Finally, Sec.~\ref{sec:mi_objective} showed that the mutual information objective is suitable for direct, gradient-based optimization. In contrast, optimizing metrics like AP and NDCG is much more challenging as they are non-differentiable,  and existing works usually resort to optimizing their surrogates \cite{lin2016structured,wang2015ranking,APSVM} rather than gradient-based optimization.
Furthermore, mutual information itself is essentially parameter-free, whereas many other hashing formulations  require (and can be sensitive to) tuning parameters, such as thresholds or margins \cite{MLH,Wang_ACCV2016}, quantization strength \cite{Li_IJCAI2016,Lin_CVPR2016,Shen_CVPR2015}, \etc.

% ------------------------ Experiments
% ------------------------ Experiments
% ------------------------ Experiments
\section{Experiments}
\label{sec:experiments}
\vspace*{-0.25em}
We evaluate our approach on three widely used image benchmarks. 
We first describe the datasets and experimental setup in Sec.~\ref{sec:setup}. We evaluate the mutual information update criterion in Sec.~\ref{sec:eval_TU} and %then evaluate 
the mutual information based objective function for learning hash mappings in Sec.~\ref{sec:eval_mihash}. 
Our implementation is publicly available at \url{https://github.com/fcakir/mihash}.

%---------------------- fig
%---------------------- fig
\begin{figure*}[t]
\centering
\includegraphics[trim=2cm 0cm 3cm .8cm,clip,width=.98\linewidth]{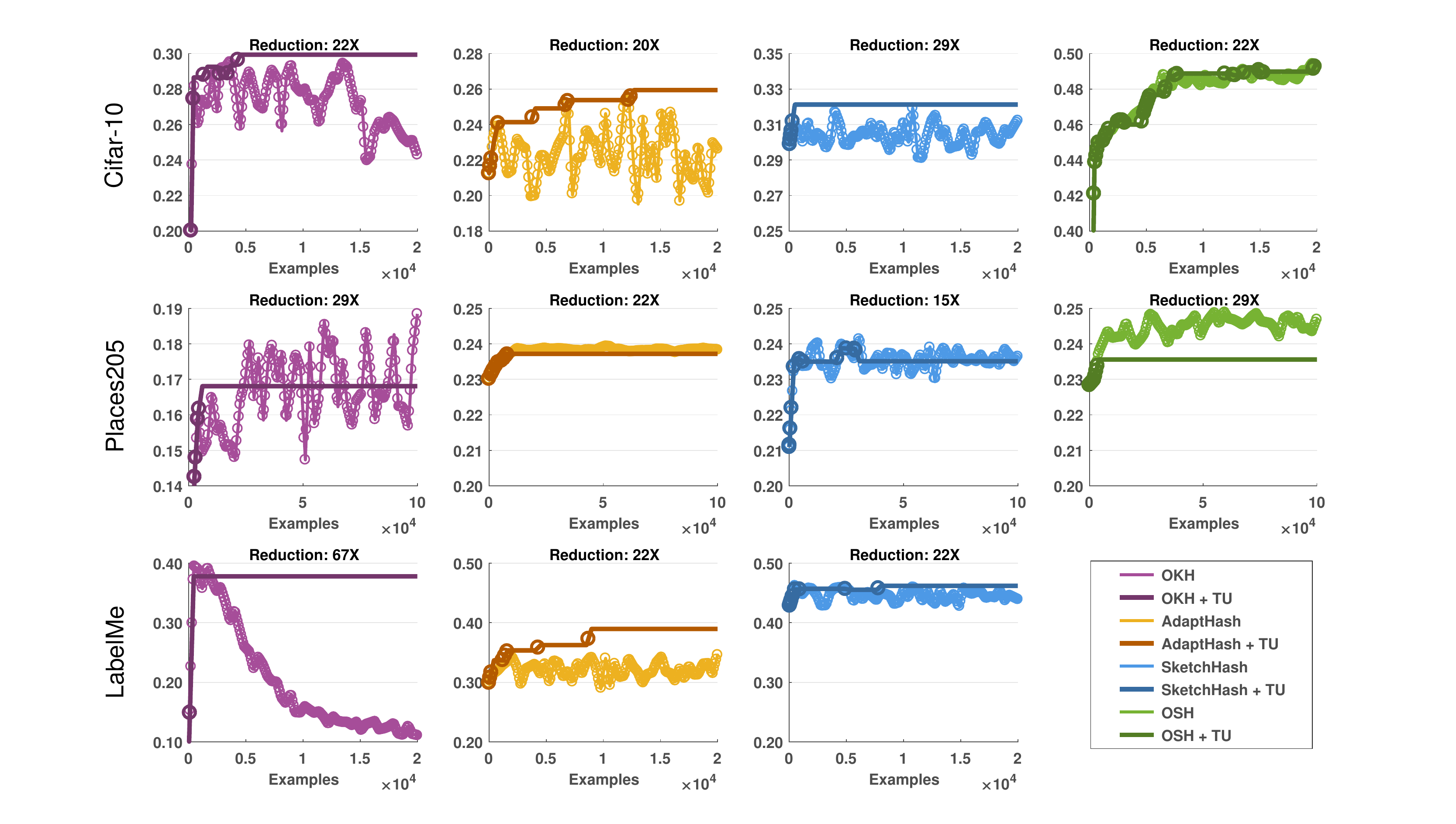}
\vspace{-1.2em}
\caption{Retrieval mAP \vs. number of processed training examples for four hashing methods on the three datasets, with and without Trigger Update (\texttt{TU}). We use default threshold $\theta=0$ for {\tt TU}. Circles indicate hash table updates, and the ratio of reduction in the number of updates is shown in the titles. {\tt TU} substantially reduces the number of updates while having a stabilizing effect on the retrieval performance.
Note: since OSH \cite{OECC} assumes supervision in terms of class labels, it is not applicable to the unsupervised LabelMe dataset.
}
\label{fig:trig_update}
\end{figure*}

%---------------------- Datasets
%---------------------- Datasets
\subsection{Datasets and Experimental Setup}
\label{sec:setup}
{\bf CIFAR-10} is a widely-used dataset for image classification and retrieval, containing 60K images from 10 different categories \cite{krizhevsky2009learning}.
For feature representation, we use CNN features extracted from the $fc7$ layer of a VGG-16 network \cite{simonyan2014very} pre-trained on ImageNet. 

{\bf Places205} is a subset of the large-scale Places dataset \cite{zhou2014learning} for scene recognition. Places205 contains 2.5M images from 205 scene categories.
This is a very challenging dataset due to its large size and number of categories, and it has not been studied in the hashing literature to our knowledge.
We extract CNN features from the $fc7$ layer of an AlexNet \cite{krizhevsky2012imagenet} pre-trained on ImageNet, and reduce the  dimensionality to 128 using PCA.

{\bf LabelMe}. The 22K LabelMe dataset \cite{russell2008labelme,torralba2008small} has 22,019 images represented as 512-dimensional GIST descriptors.
This is an unsupervised dataset without labels, and standard practice uses the Euclidean distance to determine neighbor relationships. Specifically, $\mathbf{x}_i$ and $\mathbf{x}_j$ are considered neighbor pairs if their Euclidean distance is within the smallest 5\% in the training set. For a query, the closest 5\% examples are considered true neighbors. 

All datasets are randomly split into a retrieval set and a test set, and a subset from the retrieval set is used for learning hash functions.
Specifically, for \textbf{CIFAR-10}, the test set has 1K images and the retrieval set has 59K. A random subset of 20K  images from the retrieval set is used for learning, and the size of the reservoir is set to 1K. 
For \textbf{Places205}, we sample 20 images from each class to construct a test set of 4.1K images, and use the rest as the retrieval set. A random subset of 100K images is used to for learning, and the reservoir size is 5K. 
For \textbf{LabelMe}, the dataset is split into retrieval and test sets with 20K and 2K samples, respectively. Similar to CIFAR-10, we use a reservoir of size 1K.

For online hashing experiments, we run three randomized trials for each experiment and report averaged results.
To evaluate retrieval performances, we adopt the widely-used mean Average Precision (mAP).
Due to the large size of Places205, mAP is very time-consuming to compute, and we compute mAP on the top 1000 retrieved examples (mAP@1000), as done in \cite{Lin_CVPR2016}.

%---------------------- fig
%---------------------- fig
\begin{figure*}[t]
\centering
\includegraphics[trim={6cm 0 7.5cm 0},width=.98\linewidth]{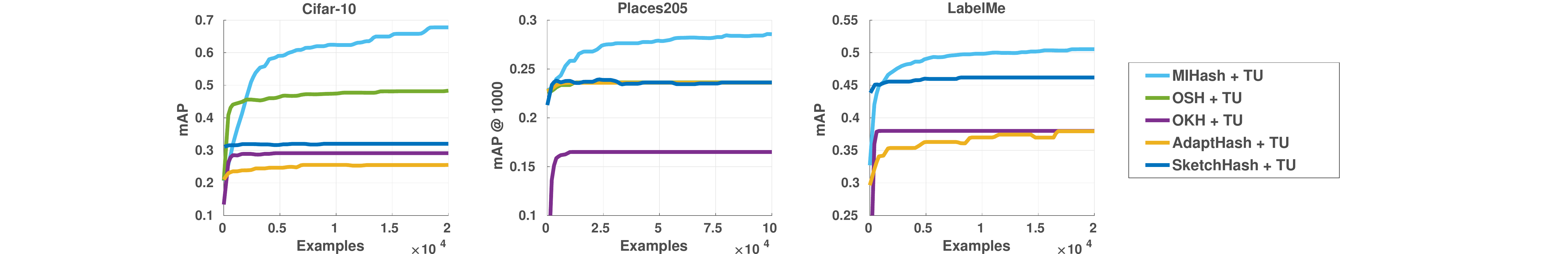}
\caption{Online hashing performance  comparison on three datasets, where all methods use the Trigger Update module ({\tt TU}) with default threshold $\theta=0$. 
\mihash~clearly outperforms other competing methods.
OSH, AdaptHash, and SketchHash perform very similarly on Places205, thus their curves overlap. 
}
\label{fig:mihash}
\end{figure*}

%----------------------Eval: Update
\subsection{Evaluation: Update Criterion}
\label{sec:eval_TU}
We evaluate our mutual information based update criterion, the Trigger Update module ({\tt TU}).
We apply {\tt TU} to all existing online hashing methods known to us: Online Kernel Hashing (OKH) \cite{OKH}, Online Supervised Hashing (OSH) \cite{OECC}, Adaptive Hashing (AdaptHash) \cite{cakir2015adaptive} and Online Sketching Hashing (SketchHash) \cite{osketchhash}. We use publicly available implementations of all methods.
The hash code length is fixed at $32$ bits.

As our work is the first in addressing the hash table update criterion for online hashing, we compare to a data-agnostic baseline, which updates the hash table at a fixed rate.
The rate is controlled by a parameter $U$, referred to as the ``update interval": 
after processing every $U$ examples, the baseline unconditionally triggers an update, while {\tt TU} makes a decision using the mutual information criterion.
For each dataset, 
$U$ is set such that the baseline updates 201 times in total.
This ensures that the baseline is never too outdated, but updates are still fairly infrequent: in all cases, the smallest $U$ is 100. 

\textbf{Results for the Trigger Update module.}
Fig.~\ref{fig:trig_update} depicts the  retrieval mAP over time for all four online hashing methods considered, on three datasets, with and without incorporating \texttt{TU}. 
We can clearly observe a significant reduction in the number of hash table updates, between one and two orders of magnitude in all cases. For example, the number of hash table updates is reduced by a factor of $67$ for the OKH method on LabelMe. 

The quality-based update criterion is particularly important for hashing methods that may yield inferior hash mappings due to noisy data and/or imperfect learning techniques.
In other words, {\tt TU} can be used to {filter} updates to the hash mapping with negative or small improvement.
This  has a stabilizing effect on the mAP curve,  notably for OKH and AdaptHash. 
For OSH, which appears to stably improve over time, {\tt TU} nevertheless significantly reduces revisits to the hash table while maintaining its performance. 

All results in Fig.~\ref{fig:trig_update} are obtained  using the default threshold parameter $\theta=0$, defined in Eq.~\ref{mi_step_criteria}. 
We do not tune $\theta$ in order to show general applicability.
We also discuss the impact of the reservoir set $\mathcal{R}$. There is a trade-off regarding the size of $\mathcal{R}$: a larger $\mathcal{R}$ leads to better approximation but increases the overhead. Nevertheless, we observed robust and consistent results with $|\mathcal{R}|$ not exceeding 5\% of the size of the training stream.

%---------------------- Eval: Objective Function
\subsection{Evaluation: Learning Objective}
\label{sec:eval_mihash}
We evaluate the mutual information based hashing objective.
We name our method \mihash,  and train it using stochastic gradient descent (SGD).
This allows it to be applied to both the online setting and batch setting in learning hash functions.

During minibatch-based SGD, to compute the mutual information objective in Eq.~\ref{eq:mi_obj} and its gradients, we need access to the sets $\vp_\xhat$, $\vm_\xhat$ for each considered $\xhat$, in order to estimate $p_\D^+$ and $p_\D^-$.
For the online setting in Sec.~\ref{sec:eval_mihash_online}, a standalone reservoir set $\mathcal{R}$ is assumed as in the previous experiment, and we partition $\mathcal{R}$ into $\{\vp_\xhat,\vm_\xhat\}$ with respect to each incoming $\xhat$.
In this case, even a batch size of 1 can be used.
For the batch setting in Sec.~\ref{sec:eval_mihash_batch}, 
$\{\vp_\xhat,\vm_\xhat\}$ are defined within the same minibatch as $\xhat$.

\vspace{-.5em}
\subsubsection{Online Setting}
\label{sec:eval_mihash_online}
 We first consider an online setting that is the same as in Sec.~\ref{sec:eval_TU}. 
We compare against other online hashing methods: OKH, OSH, AdaptHash and SketchHash. All methods are equipped with the {\tt TU} module with the default threshold $\theta=0$, which has been demonstrated to work well.% in previous experiments.

%\noindent
\textbf{Results for Online Setting.} We first show the mAP curve comparisons in Fig.~\ref{fig:mihash}. For competing online hashing methods, the curves are the same as the ones with {\tt TU} in Fig.~\ref{fig:trig_update}, and we remove markers to avoid clutter. 
\mihash~clearly outperforms other online hashing methods on all three datasets, and shows potential for further improvement with more training data.
The combination of {\tt TU} and \mihash~gives a complete online hashing system that enjoys a superior learning objective with a plug-in update criterion that improves efficiency.

\begin{figure}
\centering
\includegraphics[trim=2.2cm 0 1.3cm 0,width=0.90\linewidth]{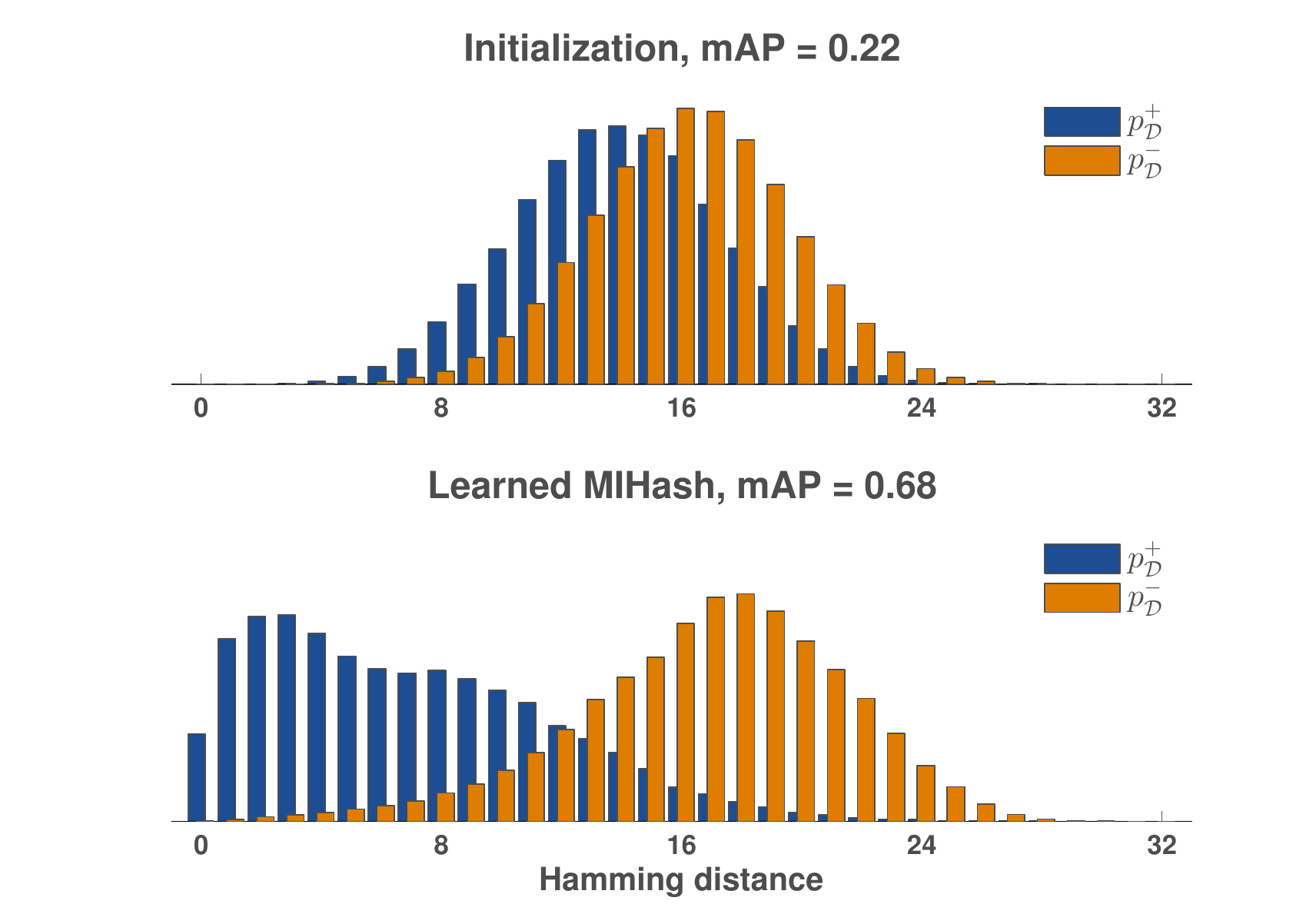}
\vspace{-.5em}
\caption{We plot the distributions $p_{\mathcal{D}}^+$ and $p_{\mathcal{D}}^-$, averaged on the CIFAR-10 test set, before and after learning \mihash~with 20K training examples. 
Optimizing the mutual information objective substantially reduces the overlap between them, resulting in state-of-the-art mAP for the online setting, as shown in  Fig.~\ref{fig:mihash}.
}
\label{fig:distance}
\end{figure}

We next give insights into the distribution-separating effect from optimizing mutual information. In Fig.~\ref{fig:distance}, we plot the conditional distributions $p_\D^+$ and $p_\D^-$ averaged on the CIFAR-10 test set, before and after learning \mihash~with the 20K training examples. Before learning, with a randomly initialized hash mapping, $p_\D^+$ and $p_\D^-$ exhibit high overlap.
After learning, \mihash~achieves good separation between $p_\D^+$ and $p_\D^-$: the overlap reduces significantly, and the mass of $p_\D^+$ is pushed towards 0. This separation is reflected in the large improvement in mAP (0.68 vs. 0.22).

In contrast with the other methods, the mutual information formulation is parameter-free.
For instance, there is no threshold parameter that requires separating $p_\D^+$ and $p_\D^-$ at a certain distance value. Likewise, there is no margin parameter that dictates the amount of separation in absolute terms.
Such parameters usually need to be tuned to fit to data, whereas the optimization of mutual information is automatically guided by the data itself.

% %---------------------- fig
% %---------------------- fig

\vspace{-0.5em}
\subsubsection{Batch Setting}
\label{sec:eval_mihash_batch}
To further demonstrate the potential of \mihash, we consider the batch learning setting, where the entire training set is available at once. 
We compare against state-of-the-art batch formulations,  including: Supervised Hashing with Kernels (SHK) \cite{SHK}, Fast Supervised Hashing with Decision Trees (FastHash) \cite{FastHash}, Supervised Discrete Hashing (SDH) \cite{Shen_CVPR2015}, Efficient Training of Very Deep Neural Networks (VDSH) \cite{Ziming_VeryDeep_CVPR2016}, Deep Supervised Hashing with Pairwise Labels (DPSH) \cite{Li_IJCAI2016} and Deep Supervised Hashing with Triplet Labels (DTSH) \cite{Wang_ACCV2016}. 
These competing methods have shown to outperform earlier and other work such as \cite{ITQ, BRE, MLH, Xia_AAAI2014, Lai_CVPR2015, Zhao_CVPR2015}.
We focus on comparisons on the CIFAR-10 dataset, which is the canonical benchmark for supervised hashing. 
Similar to \cite{Wang_ACCV2016}, we consider two experimental settings, which we detail below.

\textbf{Setting 1}: 
5K training examples are sampled for learning hash mappings, and 1K examples are used as the test set. 
All methods learn shallow models on top of \emph{fc7} features from a VGG-16 network \cite{simonyan2014very} pretrained on ImageNet. 
For three gradient-based methods (DPSH, DTSH, and \mihash), this means learning linear hash functions.
Note that VDSH uses customized architectures consisting of only fully-connected layers, and it is unclear how to adapt it to use standard architectures; we used its full model with 16 layers and 1024 nodes per layer.

\textbf{Setting 2}:
We use the full training set of size 50K and test set of size 10K.
We focus on comparing the end-to-end performance of \mihash~against two recent leading methods: DPSH and DTSH, using the same VGG-F network architecture \cite{Chatfield14} that they are trained on.
We use publicly available implementations for the compared methods, and exhaustively search parameter settings for them.
For \mihash, the minibatch size is set to 100 and 250 in Settings 1 and 2, respectively.
We use SGD with momentum, and decrease the learning rate when the training loss saturates.
See supplementary material for more details.

\textbf{Results for Batch Setting.} In Table~\ref{table:cifar_exp_soa}, we list batch learning results for all methods.
In Setting 1, \mihash~outperforms all competing methods in terms of mAP, in some cases with only a single training epoch (\eg against VDSH, DPSH).
This suggests that mutual information is a more effective learning objective for hashing. 
\mihash~learns a linear layer on the input  features, while some other methods can learn non-linear hash functions: for instance, the closest competitor, FastHash, is a two-step hashing method based on sophisticated binary code inference and boosted trees.

In Setting 2,  with end-to-end finetuning, \mihash~significantly outperforms DPSH and DTSH, the two most competitive deep hashing methods, and sets the current state-of-the-art for CIFAR-10. 
Again, note that \mihash~has no tuning parameters in its objective function. In contrast, both DPSH and DTSH have parameters to control the quantization strength that need to be tuned.

% ------------------------ table 
% ------------------------ table 
\begin{table}[!t]
\centering
\begin{threeparttable}
\small
{
\begin{tabular}{cl|v|v|v|v} %|c
\hline
& \multirow{3}{*}{\textbf{Method}}  & \multicolumn{4}{c}{\textbf{Code Length}} \\ 
\cline{3-6} 
 &  & 12 & 24 & 32 & 48 \\ 
\hline
{\multirow{8}{*}{\rotatebox[origin=c]{90}{{{Setting 1}}\hspace*{0cm}}}} 
& SHK & 0.497 & 0.615 & 0.645 & 0.682 \\
& SDH & 0.521 & 0.576 & 0.589 & 0.592 \\
& VDSH &  0.523 & 0.546 & 0.537 & 0.554 \\
& DPSH & 0.420 & 0.518 & 0.538 & 0.553 \\
& DTSH & 0.617 &  0.659 & 0.689  & 0.702 \\
& FastHash & {0.632} & {0.700}  & {0.724} & {0.738} \\
%\cdashline{2-6}
& \mihash\tnote{1} & {0.524} & {0.563} & 0.597 & 0.609  \\ 
& \mihash  & \textbf{0.683} & \textbf{0.720} & \textbf{0.727} & \textbf{0.746} \\ 
\hline
& \textbf{Method} & 16 & 24 & 32 & 48\\ \hline
{\multirow{5}{*}{\rotatebox[origin=c]{90}{{{Setting 2}}\hspace*{0cm}}}} 
& DPSH\tnote{2}  &  0.763 & 0.781 & 0.795 & 0.807 \\ 
& DTSH\tnote{2}  &  0.915 & 0.923 & 0.925 & 0.926 \\ 
& DPSH  &  0.908 & 0.909 & 0.917 & 0.932 \\ 
& DTSH  &  0.916 &  0.924 & 0.927  & 0.934 \\ 
%\cdashline{2-6}
& \mihash  & \textbf{0.929} & \textbf{0.933} & \textbf{0.938} & \textbf{0.942} \\ 
\hline
\end{tabular}
}
\begin{tablenotes}
\item [1] Results after a single training epoch.
\item[2] Results as reported in DPSH \cite{Li_IJCAI2016} and DTSH \cite{Wang_ACCV2016}.
\end{tablenotes}
\vspace{-.5em}
\caption{Comparison against state-of-the-art hashing methods on CIFAR-10.
We report mean Average Precision (mAP) on the test set, with best results in \textbf{bold}.
See text for the details of the two experimental settings.
} 
\label{table:cifar_exp_soa}
\end{threeparttable}
\end{table}

% ------------------------ conclusion
% ------------------------ conclusion
% ------------------------ conclusion
\section{Conclusion}

We advance the state-of-the-art for online hashing in two aspects. In order to resolve the issue of hash table updates in online hashing, we define a quality measure using the mutual information between variables induced by the hash mapping.
This measure is efficiently computable, correlates well with standard evaluation metrics, and leads to consistent computational savings for existing online hashing methods while maintaining their retrieval accuracy.
Inspired by these strong results, we further propose a hashing method \mihash, by optimizing mutual information as an objective with stochastic gradient descent.
In both online and batch settings, \mihash~achieves superior performance compared to state-of-the-art hashing techniques.

\section*{Acknowledgements}
This research was supported in part by a BU IGNITION award, US NSF grant 1029430, and gifts from NVIDIA.

% ------------------------ references
% ------------------------ references
% ------------------------ references
{\small
\bibliographystyle{ieee}
\bibliography{egbib}
}
\clearpage

% ----------------------- appendix
% ----------------------- appendix
% ----------------------- appendix
\section*{Appendix}
\appendix

\section{Implementation Details of \mihash}
We discuss the implementation details of \mihash. In the online hashing experiments, for simplicity we model \mihash~using linear hash functions, in the form of $\phi_i(\x)=\mathrm{sgn}(w_i^\top\x)\in\{-1,+1\},i=1,\ldots,b$. The learning capacity of such a model is lower than the kernel-based OKH, and is the same as OSH, AdaptHash, and SketchHash, which use linear hash functions as well.

For the batch hashing experiments, as mentioned in the paper, 
we similarly model \mihash~using linear hash functions in the first setting, but perform end-to-end learning with the VGG-F network in the second setting. 
In this case, the hash functions become
$\phi_i(\x)=\mathrm{sgn}(f_i(x;w))\in\{-1,+1\},i=1,\ldots,b$, where $f_i$ are the logits produced by the previous layer in the network.

We train \mihash~using stochastic gradient descent. In Eq.~11 in the paper, we gave the gradients of the mutual information objective $\mathcal{I}$ with respect to the \emph{outputs} of the hash mapping, $\Phi(\x)$. Both $\mathcal{I}$ and $\partial\mathcal{I}/\partial\Phi(\x)$ are parameter-free.
In order to further back-propagate gradients to the \emph{inputs} of $\Phi(\x)$ and model parameters $\{w_i\}$, 
we approximate the $\mathrm{sgn}$ function using the sigmoid function $\sigma$: 
\begin{align}
\phi_i(\x)\approx 2\sigma(Aw_i^\top\x)-1,
\end{align}
where $A>1$ is a scaling parameter, used to increase the ``sharpness" of the approximation. We find $A$ from the set $\{10,20,30,40,50\}$ in our experiments.

We note that $A$ is {not} a tuning parameter of the mutual information objective, but rather a parameter of the underlying hash functions. The design of the hash functions  is not coupled with the mutual information objective, thus can be separated. It will be an interesting topic to explore other methods of constructing hash functions, potentially in ways that are free of tuning parameters.

\section{Experimental Details}
\subsection{The streaming scenario}
We set up a streaming scenario in our online hashing experiments.
We run three randomized trials for each experiment. In each trial, we first randomly split the dataset into a retrieval set and a test set as described in Sec.~4.1 in the paper, and randomly sample the training subset from the retrieval set. The ordering of the  training set is  also randomly permuted. 
The random seeds are fixed, so the baselines and methods  with the Trigger Update module observe the same training sequences.

In a streaming setting, we also measure the \emph{cumulative} retrieval performance during online hashing,  as opposed to only the final results. 
To mimic real retrieval systems where queries arrive randomly, we set 50 randomized checkpoints during the online process. We first place the checkpoints with equal spacing, then add small random perturbations to their locations.
We measure the instantaneous retrieval mAP at these checkpoints to get mAP vs. time curves (\eg curves shown in Fig.~5 in the paper), and compute the area under curve (AUC). AUC gives a summary of the entire online learning process, which cannot be reflected by the final performance at the end.

\subsection{Parameters for online hashing methods} 
\label{sec:para_online}
We describe parameters used for online hashing methods in the %32-bit 
online experiments.
Some of the competing methods require parameter tuning,  therefore we sample a validation set from the training data and find the best performing parameters for each method.
The size of the validation sets are 2K, 2K and 10K for CIFAR-10, LabelMe and Places205, respectively. 
Please refer to the respective papers for the descriptions of the parameters. 
\begin{itemize}
\vspace{-0.1em}
\item \textbf{OSH}: $\eta$ is set to 0.1 for all datasets. The ECOC codebook $C$ is populated the same way as in OSH. 
\vspace{-0.25em}
\item \textbf{AdaptHash}: the tuple $(\alpha, \lambda, \eta)$ is set to $(0.9, 0.01, 0.1)$, $(0.1, 0.01, 0.001)$ and $(0.9, 0.01, 0.1)$ for CIFAR-10, LabelMe and Places205, respectively.
\vspace{-0.25em}
\item \textbf{OKH}: the tuple $(C, \alpha)$ is set to $(0.001, 0.3)$, $(0.001, 0.3)$ and $(0.0001, 0.7)$ for CIFAR-10, LabelMe and Places205, respectively.
\vspace{-0.25em}
\item \textbf{SketchHash}: the pair (sketch size, batch size) is set to $(200, 50)$, $(100, 50)$ and $(100,50)$ for CIFAR-10, LabelMe and Places205, respectively.
%\vspace{-0.5em}
\end{itemize}

\subsection{Parameters for batch hashing methods}
\label{sec:param-batch}
We use the publicly available implementations for the compared methods, and exhaustively search parameter settings, including the default parameters as provided by the authors. 
For DPSH and DTSH we found a combination that worked well for the first setting: the mini-batch size is set to the default value of 128, and the learning rate is initialized to 1 and decayed by a factor of 0.9 after every 20 epochs. 
Additionally, for DTSH, the margin parameter is set to $b/4$ where $b$ is the hash code length.
VDSH uses a heavily customized architecture with only fully-connected layers, and it is unclear how to adapt it to work with standard CNN architectures.
In this sense, VDSH is more akin to nonlinear hashing methods such as FastHash and SHK. We used the full VDSH model with 16 layers and 1024 nodes per layer, and found the default parameters to perform the best, except that we increased the number of training iterations by an order of magnitude during finetuning.  

For \mihash, in the first setting we use a batch size of 100, and run SGD with initial learning rate of 0.1 and a decay factor of 0.5 every 10 epochs, for 100 epochs. 
For the second setting where we finetune the pretrained VGG-F network, batch size is 250, learning rate is initially set to 0.001 and decayed by half every 50 epochs.

%----------------------------------------------
%----------------------------------------------
%----------------------------------------------
\begin{table}[t]
\small
\centering
{
\hfill{}
    \begin{tabular}{l|c}
 \hline
   \begin{tabular}{@{}c@{}}\textbf{Method} \end{tabular} 
      & \begin{tabular}{@{}c@{}} \textbf{Training Time (s)}\end{tabular}
            \\
\hline
\hline
    \text{OKH} & 10.8
   \\
    \text{OKH + TU}&  23.6
    \\ %[0.2em]
    \hdashline
    \text{OSH} & 97.6
    \\ 
    
    \text{OSH + TU} & 175.8
   \\ %[0.2em]
    \hdashline

    \text{AdaptHash}   & 47.8
    \\

    \text{AdaptHash + TU} &94.8
   \\ %[0.2em]
    \hdashline

    \text{SketchHash} & 68.8
   \\
    
    \text{SketchHash + TU} & 80.0
    \\ %[0.2em]
     \hline %\hline
    \end{tabular}
}
\hfill{}
%\vspace{.75em}
\caption{Online hashing: running times on the CIFAR-10 20k training set, with 32-bit hash codes. For methods with the TU plugin, the added time is due to maintaining the reservoir set and computing the mutual information update criterion, and is dominated by the maintaining of the reservoir set.}
  \label{table:time}
\end{table}

\section{Running Time}
\subsection{Online Setting: Trigger Update Module}
In Table~\ref{table:time} we report running time for all methods on the CIFAR-10 dataset with 20k training examples, including time spent in learning hash functions and the added processing time for maintaining the reservoir set and computing {\tt TU}.  Numbers are recorded on a 2.3GHz Intel Xeon E5-2650 CPU workstation with 128GB of DDR3 RAM.   
Most of the added time is due to maintaining the reservoir set, which is invoked in each training iteration;  the mutual information update criterion is only checked after processing every $U=100$ examples. 
Methods with small batch sizes (\eg OSH, batch size 1) therefore incur more overhead than methods with larger batches (\eg SketchHash, batch size 50).
Results for other datasets are similar.

We note that in a real retrieval system with large-scale data, the bottleneck likely lies in recomputing the hash tables for indexed data, due to various factors such as scheduling and disk I/O. We reduce this bottleneck significantly by using {\tt TU}. 
Compared to this bottleneck, the increase in training time is not significant.

\subsection{Batch Setting}
Table \ref{table:cifar_exp_time} reports CPU times for learning 48-bit hash mappings in the first experimental setting on CIFAR-10 (5K training set).
Retrieval mAP are replicated from Table~1 in the paper.
For learning a single layer, our Matlab implementation of \mihash~achieves 1.9 seconds per epoch on CPU. 
\mihash~achieves competitive performance with a single  epoch,  and has a total training time on par with FastHash, while yielding superior performance.

% ------------------------ table 
% ------------------------ table 
\begin{table}[!t]
\small
\centering
{
%\hfill{}
\begin{tabular}{l|c|c} %|c
\hline
\textbf{Method} & \textbf{mAP} & \textbf{Training Time (s)}\\
\hline
\hline
SHK  & 0.682 &  180 \\
SDH  & 0.592 & 4.8  \\
FastHash   & 0.738 & 140 \\
VDSH*  & 0.554 & 206  \\
DPSH  & 0.553 & 450  \\
DTSH  & 0.702 & 1728  \\
\hdashline
\mihash, 1ep & 0.609 & 1.9 \\
\mihash  & 0.746 & 190  \\
\hline
\end{tabular}
}
%\vspace{.75em}
\caption{Batch hashing: test performance and training time for 48-bit codes on the CIFAR-10, using the 5k training set. *VDSH is trained with the full model as detailed in \ref{sec:param-batch}. {1ep} stands for training for one epoch only.} 
  \label{table:cifar_exp_time}
\end{table}

\afterpage{
%---------------------- fig
%---------------------- fig
\begin{figure*}[t]
\centering
\includegraphics[trim={2.5cm 0 3cm 0},width=.98\linewidth]{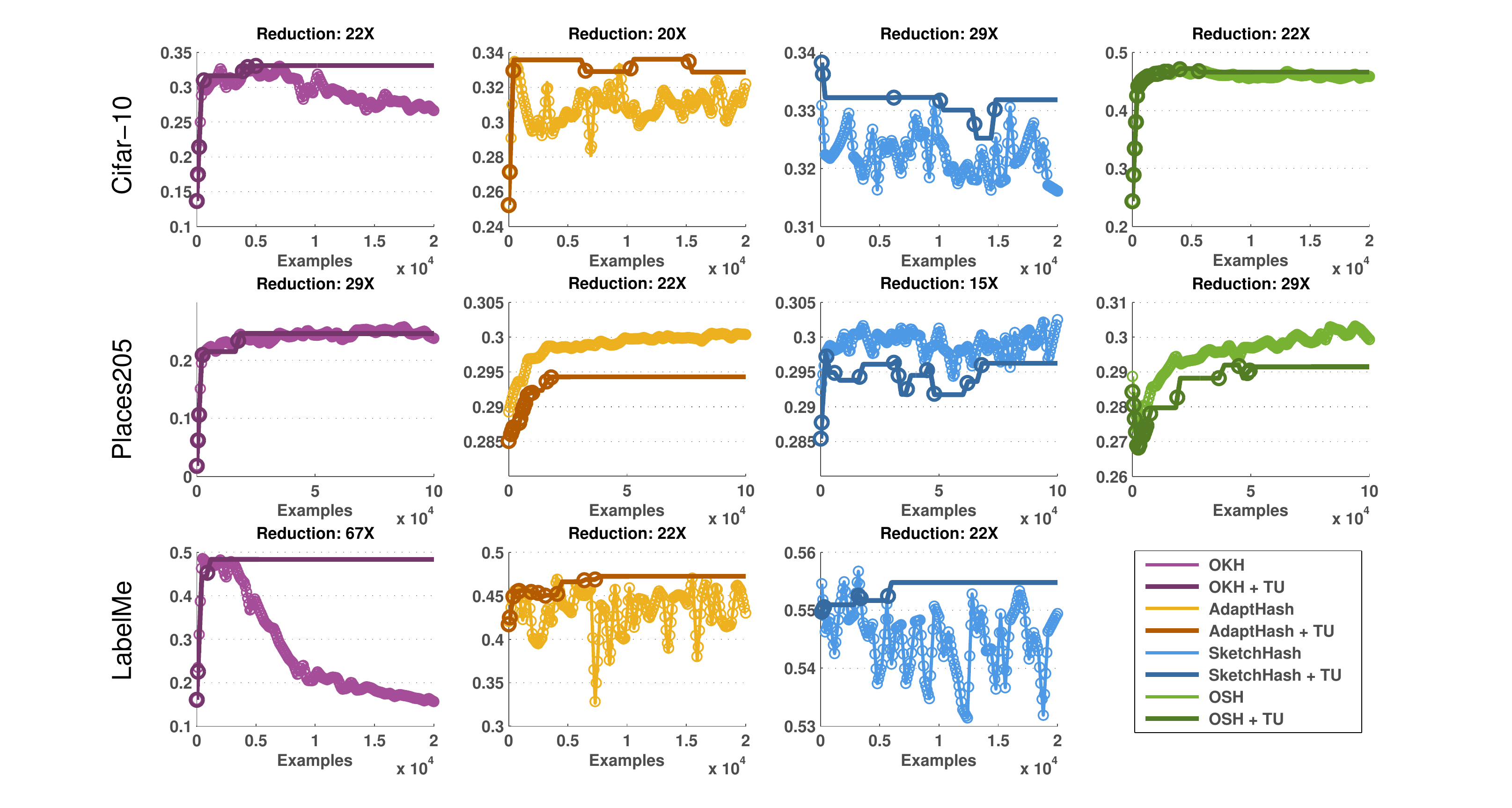}
\caption{64-bit experiments: Retrieval mAP \vs. number of training examples for four existing online hashing methods on the three datasets, with and without Trigger Update (\texttt{TU}). We use default threshold $\theta=0$ for {\tt TU}. Circles indicate hash table updates, and the ratio of reduction in the number of updates is marked for each graph. {\tt TU} substantially reduces the number of updates while having a stabilizing effect on the retrieval performance.
Note: since the OSH method  assumes supervision in terms of class labels, it is not applicable to the unsupervised LabelMe dataset.
}
\label{fig:TU64}
\end{figure*}

%---------------------- fig
%---------------------- fig
\begin{figure*}[t]
\centering
\includegraphics[trim={6cm 0 8.5cm 0},width=0.95\linewidth]{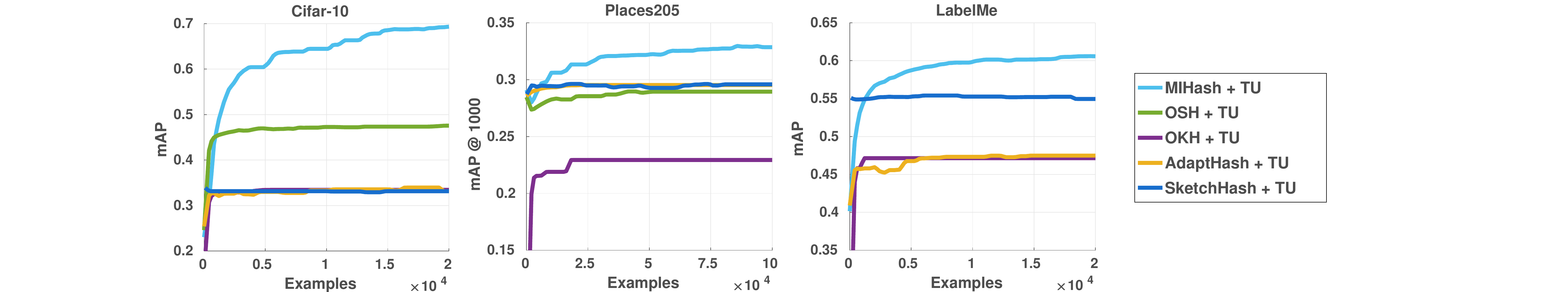}
\vspace{1em}
\caption{64-bit experiments: Online hashing performance (mAP) comparison on three datasets, where all methods use the Trigger Update module ({\tt TU}) with $\theta=0$. Using the mutual information objective, \mihash~clearly outperforms other methods.
OKH, AdaptHash, ad SketchHash perform very similarly on CIFAR-10. OSH, AdaptHash, ad SketchHash perform very similarly on Places205. 
Again, the OSH method is not applicable to the unsupervised LabelMe dataset.
}
\label{fig:mihash64}
\end{figure*}

\clearpage
}

%----------------------------------------------
%----------------------------------------------
%----------------------------------------------

\section{Additional Experimental Results}
\subsection{Online Hashing: Other Code Lengths}
In the online hashing experiments  we reported in the paper, all online hashing methods are compared in  the same setup with 32-bit hash codes. Additionally,  we also present results using  64-bit hash codes on all three datasets. 
The parameters for all methods are found through validation as described in \ref{sec:para_online}.

Similar to Sec~4.2 in the paper, we show the comparisons  with  and without {\tt TU} for existing online hashing methods in Fig.~\ref{fig:TU64}, and plot the mAP curves for all methods, including \mihash, in Fig.~\ref{fig:mihash64}.  The 64-bit results are uniformly better than 32-bit results for all methods in terms of mAP, but still follow the same patterns.
Again, we can see that \mihash~clearly outperforms all competing online hashing methods, and shows potential for improvement given more training data.

% ------------ theta
% ------------ theta
\subsection{Parameter Study: $\theta$}
We present a parameter study on the parameter $\theta$, the improvement threshold on the mutual information criterion in {\tt TU}.
In our previous experiments, we found the default $\theta=0$ to work well, and did not specifically tune $\theta$.
However, tuning for a larger $\theta$ could lead to better trade-offs, since small improvements in the quality of the hash mapping may not justify the cost of a full hash table update.

For this study, we vary parameter $\theta$ from $-\infty$ to $\infty$ for all methods (with 32-bit hash codes). 
$\theta=-\infty$ reduces to the baseline. On the other hand, $\theta=\infty$ prevents any updates to the initial hash mapping and hash table, and results in only one hash table update (for the initial mapping) and typically low performance.
The performance metric  we focus on in this study is the cumulative metric, AUC, since it better summarizes the entire online learning process than the final performance alone.

We use a custom update schedule for SketchHash:  we enforce hash table updates in the early iterations regardless of other criteria, until the number of observed examples reaches the specified size of the ``data sketch", which SketchHash uses to perform a batch hashing algorithm. This was observed to be critical for the performance of SketchHash. Therefore, the number of hash table updates for SketchHash can be greater than  1 even for $\theta=\infty$.

We present full results in Tables~\ref{table:trigger_update_cifar}, \ref{table:trigger_update_places}, \ref{table:trigger_update_labelme}.
In all cases, we observe a substantial decrease in the number of hash table updates as $\theta$ increases. With reasonable $\theta$ values (typically around 0), the number of hash table updates can be reduced by over an order of magnitude with no loss in AUC. Note that the computation-performance trade-off achieved by the default $\theta=0$ is always among the best, thereby in practice it can be used  without tuning.

% ------------ U
% ------------ U
\subsection{Parameter  Study: $U$}
We  simulate a data-agnostic baseline that updates hash tables at a constant rate, using the update interval parameter $U$.
In the paper, $U$ is set such that the baseline updates a total of 201 times for all datasets. 
This ensures that the baseline is never too outdated (compared to 50 checkpoints at which performance is evaluated), but is still fairly infrequent: the smallest $U$ in this case is 100, which means the baselines process at least 100 training examples before recomputing the hash table.
For completeness, here we present the results using different values of $U$, where all methods again use 32-bit hash codes and the default $\theta=0$. 

We used a simple rule that avoids unnecessary hash table updates if the hash mapping itself does not change. Specifically, we do not update if $\|\Phi_t-\Phi^s\|<10^{-6}$, where $\Phi^s$ is the current snapshot and $\Phi_t$ is the new candidate. Some {baseline} entries have fewer updates because of this rule (\eg AdaptHash on Places205).
And as explained before, due to the custom update schedule, SketchHash may have more hash table updates than what is suggested by $U$.

Please see Tables~\ref{table:u_cifar}, \ref{table:u_labelme}, \ref{table:u_places} for the full results.
In all experiments, we run three random trials and average the results as mentioned before, and the standard deviation of mAP and AUC scores are less than $0.01$.
Generally, using smaller $U$ leads to more updates by both the baselines and methods with {\tt TU}; recall that $U$ is also a parameter of {\tt TU} which specifies the frequency of checking the update criterion.
However, methods with the {\tt TU} module appear to be quite insensitive to the choice of $U$, \eg the number of updates for SketchHash with {\tt TU} on CIFAR-10 only increases by 2x while $U$ is reduced by 20x, from 1000 to 50. We attribute this to the ability of {\tt TU} to filter out unnecessary updates. 
Across different values of $U$, {\tt TU} consistently brings computational savings while preserving/improving online hashing performance, as indicated by final mAP and AUC.

% -----------------------------------------------
% -----------------------------------------------

% ---------------------------------------------------------
% ------------------------ Cifar
\begin{table}[t]
\small
\centering
{
    \begin{tabular}{c|c|c|v}
 \hline
 \multicolumn{4}{c}{\makecell{\textbf{CIFAR-10, 32 bits}}} \\
 \hline
    %-------------- Cifar OKH 
   \begin{tabular}{@{}c@{}} OKH \end{tabular} 
   & \begin{tabular}{@{}c@{}} \textbf{HT Updates} \end{tabular}
   & \begin{tabular}{@{}c@{}}\textbf{AUC} \end{tabular}
   & \begin{tabular}{@{}c@{}}$\Delta$\textbf{AUC} \end{tabular} \\
\hline
    $\leq -0.1$  & $201$  & $0.259$ & -- \\
    $-0.01$ & $190$ ($5.8$x) & $0.260$ & $+0.4\%$ \\ 
    $-10^{-4}$ & $8.0$ ($25.1$x) &  $0.287$ & $+10.8\%$ \\ 
	$0$ & $8.0$ ($25.1$x) & $0.287$ & $+10.8\%$ \\
    $10^{-4}$ & $7.7$ ($26.1$x) &  0.287 & $+10.8\%$ \\
    $0.01$   & $3.3$ ($91.2$x) &  0.280 & $+8.1\%$ \\
    $\geq 0.2$  & $1.0$ ($201$x) & 0.134 & $-48.3\%$ \\
    \hline %\hline 
    
 %-------------- Cifar OSH
\begin{tabular}{@{}c@{}} OSH \end{tabular}  
& \begin{tabular}{@{}c@{}} \textbf{HT Updates} \end{tabular} 
& \begin{tabular}{@{}c@{}}\textbf{AUC} \end{tabular}
& \begin{tabular}{@{}c@{}}$\Delta$\textbf{AUC} \end{tabular} 
\\
\hline
    $\leq -0.01$  & $201$   & $0.463$ &  --\\ % \pm 0.01 $ \\
    $- 10^{-4}$  &  $39.0$ ($5.2\text{x}$)  & $0.466$  & ${+0.6\%}$\\    
	$ 0$ & $36.7$ ($5.5\text{x}$) & $0.466$  & $+0.6\%$  \\
    $  10^{-4}$  &  $35.7$ ($5.6\text{x}$)	 & $0.466$ & $+0.6\%$ \\
    $ 0.01$ &  $6.7$  ($30\text{x}$)  & $0.453$  & $-2.1\%$  \\
    $ 0.1$  & $2.0$  ($100\text{x}$)  & $0.386$  & $-16\%$ \\
    $ \geq 0.3 $ & $1.0$ ($201\text{x}$) & $0.207$  & $-55\%$ \\
     \hline %\hline

    %-------------- Cifar AdaptHash 
   \begin{tabular}{@{}c@{}} AdaptHash \end{tabular} 
   & \begin{tabular}{@{}c@{}} \textbf{HT Updates} \end{tabular}
   & \begin{tabular}{@{}c@{}} \textbf{AUC} \end{tabular}
   & \begin{tabular}{@{}c@{}} $\Delta$\textbf{AUC} \end{tabular} \\
  \hline
    $\leq -0.1$  & $201$  & $0.218$ & -- \\
    $-0.01$ & $68.3$ ($2.9$x) & $0.238$ & $+9.2\%$ \\ 
    $-10^{-4}$ & $10.3$ ($19.5$x) & $0.250$ & $+14.7\%$ \\
	$0$ & $10.0$ ($20.1$x) & $0.250$ & $+14.7\%$ \\ 
    $10^{-4}$ & $10.0$ ($20.1$x) & $0.250$ & $+14.7\%$ \\
    $0.01$ & $3.3$ ($60.9$x) & $0.244$ & $+11.9\%$ \\
    $\geq 0.1$  & $1.0$ ($201$x) & $0.211$ & $-3.3\%$ \\
 \hline %\hline 

    %-------------- Cifar SketchHash 
   \begin{tabular}{@{}c@{}} SketchHash \end{tabular} 
   & \begin{tabular}{@{}c@{}} \textbf{HT updates} \end{tabular}
   & \begin{tabular}{@{}c@{}}\textbf{AUC} \end{tabular}
   & \begin{tabular}{@{}c@{}}$\Delta$\textbf{AUC} \end{tabular} \\
\hline
    $\leq -0.01$  & $201$  & $0.304$ & -- \\
    $-10^{-4}$ & $9.0$ ($22.3$x) & $0.318$ & $+4.6\%$ \\ 
    $-10^{-6}$ & $7.3$ ($27.5$x) & $0.319$ & $+4.9\%$ \\ 
	$0$ & $7.3$ ($27.5$x) &  $0.319$ & $+4.9\%$  \\
    $10^{-4}$ & $7.3$ ($27.5$x) & $0.319$ & $+4.9\%$ \\ 
    $0.01$   & $4.3$ ($46.7$x) &  $0.318$ & $+4.6\%$ \\
    $\geq0.1$   & $4.0$ ($50.3$x) & $0.314$ & $+3.3\%$ \\
    \hline %\hline 
    \end{tabular}
}
\vspace{1em}
\caption{
Parameter study  on the threshold value $\theta$ for online hashing methods on \textbf{CIFAR-10} (32 bits).
We report the number of hash table updates, 
where 100x indicates a 100 times reduction with respect to the baseline.
We also report the area under the mAP curve (AUC) and compare to baseline.
} 
  \label{table:trigger_update_cifar}
\end{table}

% ------------------------ Places
% ------------------------ Places
% ------------------------ Places
\begin{table}[t]
\small
\centering
{
    \begin{tabular}{c|c|c|v}
 \hline
     \multicolumn{4}{c}{\makecell{\textbf{Places205, 32 bits} }} \\
 \hline
    %-------------- Places OKH 
   \begin{tabular}{@{}c@{}} OKH  \end{tabular} 
   & \begin{tabular}{@{}c@{}} \textbf{HT Updates} \end{tabular} 
   & \begin{tabular}{@{}c@{}}\textbf{AUC} \end{tabular}
   & \begin{tabular}{@{}c@{}}$\Delta$\textbf{AUC} \end{tabular} 
   \\
\hline
    $\leq -0.01$  & $201$  & $0.163$ & -- \\
    $-10^{-4}$ & $8.3$ ($24.2$x) & $0.161$ & $-1.2\%$ \\ 
    $-10^{-6}$ & $7.0$ ($28.7$x) & $0.161$ & $-1.2\%$ \\ 
	$0$ & $7.0$ ($28.7$x) & $0.161$ & $-1.2\%$ \\ 
    $10^{-6}$ & $7.0$ ($28.7$x) & $0.161$ & $-1.2\%$ \\ 
    $10^{-4}$ & $5.7$ ($35.3$x) & $0.161$ & $-1.2\%$ \\ 
    $0.01$ & $2.0$ ($100$x) & $0.123$ & $-25\%$ \\
    $ \geq 0.1$  & $1.0$ ($201$x) & $0.014$ & $-91\%$ \\
    \hline %\hline 

    %-------------- Places OSH
   \begin{tabular}{@{}c@{}} OSH \end{tabular} 
   & \begin{tabular}{@{}c@{}} \textbf{HT Updates} \end{tabular}
   & \begin{tabular}{@{}c@{}}\textbf{AUC} \end{tabular}
   & \begin{tabular}{@{}c@{}}$\Delta$\textbf{AUC} \end{tabular} \\
\hline
    $\leq -0.001$  & $201$  & $0.246$ & -- \\
    $-20^{-4}$ & $101$  ($2.0$x) & $0.246$  & $0\%$ \\ 
    $-10^{-4}$ & $9.3$ ($21.6$x) & $0.236$ & $-4.1\%$ \\ 
	$0$ & $7.0$ ($28.7$x) & $0.236$ & $-4.1\%$  \\
    $10^{-4}$ & $5.7$ ($35.3$x) & $0.230$ & $-6.5\%$  \\
    $10^{-3}$ & $2.7$ ($74.4$x) & $0.224$ & $-8.9\%$  \\
    $\geq 0.1$   & $1.0$ ($201$x) & $0.226$ & $-8.1\%$   \\
    \hline %\hline 

    %-------------- Places AdaptHash 
   \begin{tabular}{@{}c@{}} AdaptHash \end{tabular} 
   & \begin{tabular}{@{}c@{}} \textbf{HT Updates} \end{tabular}
   & \begin{tabular}{@{}c@{}}\textbf{AUC} \end{tabular}
   & \begin{tabular}{@{}c@{}}$\Delta$\textbf{AUC} \end{tabular} \\
\hline
    $\leq -0.01$  & $199.7$ & $0.237$ & -- \\
    $- 10^{-4}$ & $199$ ($1.0$x) & $ 0.237$ & $0\%$ \\ 
    $-10^{-6}$ & $9.7$  ($20.6$x) & $0.236$ & $-0.4\%$ \\    
	$ 0$ & $8.7$ ($23.0$x) & $0.236$ & $-0.4\%$ \\
    $10^{-6}$ & $8.7$ ($23.0$x) & $0.235$ & $-0.8\%$  \\
    $10^{-4}$ & $3.0$  ($66.6$x) & $0.235$ & $-0.8\%$ \\
    $ \geq 0.01$  & $1.0$ ($201$x) & $ 0.227$ & $-3.4\%$ \\
    \hline %\hline 

    %-------------- Places SketchHash 
   \begin{tabular}{@{}c@{}} SketchHash \end{tabular} 
   & \begin{tabular}{@{}c@{}} \textbf{HT Updates} \end{tabular}
   & \begin{tabular}{@{}c@{}}\textbf{AUC} \end{tabular}
   & \begin{tabular}{@{}c@{}}$\Delta$\textbf{AUC} \end{tabular} \\
\hline
    $\leq -0.01$  & $201$  & $0.237$ & --  \\
    $-10^{-4}$ & $52.3$  ($3.8$x) & $0.238$ & $+0.4\%$ \\
    $-10^{-6}$ & $15.3$  ($12.6$x) & $ 0.238$ & $+0.4\%$ \\
	$0$ & $12.7$  ($15.8$x) & $0.236$ & $-0.4\%$ \\ 
    $10^{-6}$ & $15.3$  ($13.1$x) & $ 0.238$ & $+0.4\%$ \\
    $10^{-4}$   & $7.0$  ($28.7$x) & $0.239$ & $+0.8\%$ \\
    $\geq 0.01$   & $2.0$ ($101$x) & $0.223$ & $-5.9\%$ \\
    \hline %\hline 

    \end{tabular}
}
\vspace{1em}
\caption{
Parameter study  on the threshold value $\theta$ for online hashing methods on \textbf{Places205} (32 bits).
We report the number of hash table updates,  where 100x indicates a 100 times reduction with respect to the baseline.
We also report the area under the mAP curve (AUC) and compare to baseline.
} 
  \label{table:trigger_update_places}
\end{table}

% ------------------------ LabelMe
% ------------------------ LabelMe
% ------------------------ LabelMe
\begin{table}%[t]
\small
\centering
{
    \begin{tabular}{c|c|c|v}
 \hline
     \multicolumn{4}{c}{\makecell{\textbf{LabelMe, 32 bits} }} \\
 \hline
    %-------------- LabelMe OKH 
   \begin{tabular}{@{}c@{}} OKH \end{tabular} 
   & \begin{tabular}{@{}c@{}} \textbf{HT Updates} \end{tabular}
   & \begin{tabular}{@{}c@{}}\textbf{AUC} \end{tabular}
   & \begin{tabular}{@{}c@{}}$\Delta$\textbf{AUC} \end{tabular} \\
\hline
    $\leq -0.2$  & $201$  & $0.198$ & -- \\
    $-0.1$ & $196$ ($1.0$x) & $0.199$ & $+0.5\%$ \\ 
    $-0.01$ & $2.7$ ($74.4$x) & $0.373$ & $+88\%$ \\
    $-10^{-6}$ & $2.3$ ($87.4$x) & $0.374$ & $+89\%$ \\
	$0$ & $2.3$ ($87.4$x) & $0.374$ & $+89\%$ \\
    $10^{-6}$ & $2.3$ ($87.4$x) & $0.374$ & $+89\%$ \\
    $0.01$ & $2.0$ ($101$x) & $0.372$ & $+88\%$ \\
    $\geq 0.6$  & $1.0$ ($201$x) & $0.111$ & $-44\%$ \\
    \hline %\hline 

    %-------------- LabelMe AdaptHash 
   \begin{tabular}{@{}c@{}} AdaptHash \end{tabular} 
   & \begin{tabular}{@{}c@{}} \textbf{HT Updates} \end{tabular}
   & \begin{tabular}{@{}c@{}}\textbf{AUC} \end{tabular}
   & \begin{tabular}{@{}c@{}}$\Delta$\textbf{AUC} \end{tabular} \\
\hline
    $\leq -0.1$  & $201$  & $0.333$ & -- \\
    $-10^{-6}$ & $149$ ($1.3$x) & $0.330$ & $-0.9\%$ \\ 
    $-10^{-4}$ & $9.3$ ($21.6$x) & $0.365$ & $+9.6\%$ \\ 
    $-10^{-2}$ & $8.7$ ($23.1$x) & $0.365$ & $+9.6\%$ \\ 
	$0$ & $5.3$ ($37.9$x) & $0.369$ & $+11\%$  \\
    $10^{-6}$   & $8.7$ ($23.1$x) & $0.365$ & $+9.6\%$ \\
    $10^{-4}$   & $8.3$ ($24.2$x) & $0.358$ & $+7.5\%$ \\
    $10^{-2}$   & $2.7$ ($74.4$x) & $0.351$ & $+5.4\%$ \\
    $ \geq 0.1$  & $1$ ($201$x) & $0.296$ & $-11\%$  \\
    \hline %\hline 

    %-------------- LabelMe SketchHash 
   \begin{tabular}{@{}c@{}} SketchHash
   \end{tabular} & \begin{tabular}{@{}c@{}} \textbf{HT Updates} \end{tabular}
   & \begin{tabular}{@{}c@{}}\textbf{AUC} \end{tabular}
   & \begin{tabular}{@{}c@{}}$\Delta$\textbf{AUC} \end{tabular} \\
\hline
    $\leq -0.1$  & $201$  & $0.446$ & -- \\
    $-10^{-2}$ & $195$ ($1.0$x) & $0.446$ & $0\%$ \\ 
    $-10^{-4}$ & $9.3$ ($21.6$x) & $0.460$ & $+3.1\%$ \\ 
	$0$ & $8.7$ ($23.1$x) & $0.460$ & $+3.1\%$ \\
    $10^{-4}$ & $10$ ($20.1$x) & $0.459$ & $+2.9\%$ \\
    $10^{-2}$ & $4.7$ ($42.8$x) & $0.446$ & $0\%$ \\
    $ \geq 0.1$  & $4.0$ ($50.3$x) & $0.439$ & $-1.6\%$ \\
    \hline %\hline 
    \end{tabular}
}
\vspace{1em}
\caption{
Parameter study  on the threshold value $\theta$ for online hashing methods on \textbf{LabelMe} (32 bits).
We report the number of hash table updates,  where 100x indicates a 100 times reduction with respect to the baseline.
We also report the area under the mAP curve (AUC) and compare to baseline.
Note: OSH is not applicable to this unlabeled dataset since it needs supervision in terms of class labels.
} 
  \label{table:trigger_update_labelme}
\end{table}

%%%%%%%%%%%%%%%%%%%%%%%%%%%%%%%%%%%%%% CIFAR %%%%%%%%%%%%%%%%%%%%%%%%%%%%%%%%%%%%%%%%%%%
%%%%%%%%%%%%%%%%%%%%%%%%%%%%%%%%%%%%%% CIFAR %%%%%%%%%%%%%%%%%%%%%%%%%%%%%%%%%%%%%%%%%%%
\begin{table*}[!ht]
%\footnotesize
\small
\centering
{
\hfill{}
\begin{tabular}{l|c|c|c|l}
\hline
\multicolumn{5}{c}{\makecell{\textbf{CIFAR-10, 32 bits}}} \\
\hline
\begin{tabular}{@{}c@{}} \textbf{Method} \end{tabular}
& \begin{tabular}{@{}c@{}} \textbf{TU} \end{tabular}
& \begin{tabular}{@{}c@{}} \textbf{HT Updates} \end{tabular}
& \begin{tabular}{@{}c@{}}\textbf{Final mAP} \end{tabular} 
& \begin{tabular}{@{}c@{}}\textbf{AUC (mAP)} \end{tabular} 
\\
\hline
    \multirow{2}{*}{OKH, $U=10$}  & $\times$ & 1870 & 0.238  &  0.259 \\ 
    
   &   \checkmark  & 15.6 (119.3x)  & 0.297 & 0.293 (+13\%) \\ 
    
    \hdashline

    \multirow{2}{*}{OKH, $U=100$} & $\times$ & 201  & 0.238   & 0.259 \\
    
     &  \checkmark &  8 (25.1x) & 0.291 & 0.287 (+10.8\%) \\
    
    \hdashline

     \multirow{2}{*}{OKH, $U=1000$} &  $\times$ &  21  & 0.238  & 0.255  \\ 
    
     &  \checkmark & 2.6 (8x) & 0.282 & 0.273 (+7\%) \\ 

	\hline
    
    % OSH
    \multirow{2}{*}{OSH, $U=10$} &  $\times$  & 2001  & 0.480 & 0.463  \\ 
    
   & \checkmark & 110.7 (18x)  &  0.483  & 0.466 (+0.6\%) \\ 
    
    \hdashline

     \multirow{2}{*}{OSH, $U=100$} &  $\times$  & 201   & 0.480 &  0.463  \\ 
    
    &  \checkmark & 36.7 ({5.4}x) & 0.483 & 0.466 (+0.6\%) \\
    
    \hdashline

    \multirow{2}{*}{OSH, $U=1000$} & $\times$   & 21  & 0.480 & 0.454 \\ 
    
    &  \checkmark & 11.3 ({1.9}x)  & 0.479 & 0.454 \\ 
    
     %\multicolumn{5}{c}{\makecell{AdaptHash}} \\
	 \hline

    \multirow{2}{*}{AdaptHash, $U=10$} &  $\times$   & 2001 & 0.244  & 0.224   \\ 
    
     &  \checkmark & 19.6 (101.7x) & 0.267 & 0.261 (+16\%) \\ 
    
    \hdashline

    \multirow{2}{*}{AdaptHash, $U=100$} & $\times$   & 201  & 0.244  & 0.224 \\

     & \checkmark  & 10.0 (10.1x)  & 0.255 & 0.250 (+11.6\%)  \\
	
    \hdashline

    \multirow{2}{*}{AdaptHash, $U=1000$} & $\times$  &  21  & 0.244  & 0.222   \\ 
    
     & \checkmark & 5 (4.2x) & 0.252 & 0.234 (+5\%) \\ 
    
    \hline
    % \multicolumn{5}{c}{\makecell{SketchHash}} \\
    
    \multirow{2}{*}{SketchHash, $U=50$} &  $\times$  & 400  & 0.306  & 0.303  \\ 
    
     & \checkmark & 9 (44.4x) & 0.318 & 0.318 (+5\%) \\ 
    
    \hdashline

    \multirow{2}{*}{SketchHash, $U=100$} &  $\times$  & 202  & 0.306  & 0.304 \\
    
    & \checkmark  & 7.3 (27.5x)  & 0.320 & 0.319 (+4.9\%)  \\
    
    \hdashline

    \multirow{2}{*}{SketchHash, $U=1000$} & $\times$   & 24  & 0.306  & 0.305  \\ 
    
    & \checkmark & 4.6 (5.2x) & 0.317 & 0.314 (+2.9\%) \\ 
    
    \hline
    \end{tabular}
}
\hfill{}
\vspace{1em}
\caption{
Online hashing results (32 bits) with different update interval parameters ($U$) on the \textbf{CIFAR-10} dataset. 
All results are averaged from 3 random trials.
For the number of hash table updates, we report the reduction ratio (\eg 8x) for TU.
For  AUC, we report the relative change compared to baseline.
Note: SketchHash uses a batch size of 50, therefore the smallest $U$ is set to 50.
}
\vspace{1em}
\label{table:u_cifar}
\end{table*}

%%%%%%%%%%%%%%%%%%%%%%%%%%%%%%%%%%%% LABELME %%%%%%%%%%%%%%%%%%%%%%%%%%%%%%%%%%%%%%%%%%%%%%%%
\begin{table*}[!ht]
\small
  \centering
{
\hfill{}
    \begin{tabular}{l|c|c|c|l}
\hline %\hline
\multicolumn{5}{c}{\makecell{\textbf{LabelMe, 32 bits}}} \\
\hline
\begin{tabular}{@{}c@{}}\textbf{Method} \end{tabular} 
& \begin{tabular}{@{}c@{}}\textbf{TU} \end{tabular} 
& \begin{tabular}{@{}c@{}} \textbf{HT Updates} \end{tabular}
& \begin{tabular}{@{}c@{}} \textbf{Final mAP} \end{tabular} 
& \begin{tabular}{@{}c@{}}\textbf{AUC (mAP)} \end{tabular} 
\\
\hline
    % \multicolumn{5}{c}{\makecell{OKH}} \\
 \multirow{2}{*}   {OKH, $U=10$} & $\times$ &2001  & 0.119 & 0.200 \\ 
    
 & \checkmark    & 8  (250x) & 0.382 & 0.377 (+88.5\%) \\ 
    
    \hdashline

 \multirow{2}{*}{OKH, $U=100$} &$\times$ & 201  & 0.119 & 0.200\\
    
 & \checkmark     & 2.3 (86.2x) & 0.380 & 0.374 ({+87\%}) \\
    
      \hdashline

 \multirow{2}{*}   {OKH, $U=1000$} &$\times$ & 21  &  0.119 &  0.193 \\ 
    
 & \checkmark    & 2  (10.5x) & 0.373 & 0.357 (+85\%) \\ 
    
    \hline
    % AdaptHash
 \multirow{2}{*}   {AdaptHash, $U=10$} &$\times$ & 2001  & 0.318 & 0.319 \\ 
    
 & \checkmark    & 12.6 (157.9x) & 0.380  & 0.371  (+16.3\%) \\ 
    \hdashline

 \multirow{2}{*}    {AdaptHash, $U=100$} & $\times$ &201 & 0.318  & 0.318 \\
    
 & \checkmark    & 8.6 (23.1x) & 0.379 &0.365  (+14.7\%) \\
    
    \hdashline

 \multirow{2}{*}   {AdaptHash, $U=1000$} &$\times$ & 21  &  0.318 & 0.317  \\ 
    
 & \checkmark    & 5  (4.2x) & 0.343 & 0.337  (+6.3\%) \\ 
    
    \hline
    
    % \multicolumn{5}{c}{\makecell{SketchHash}} \\
 \multirow{2}{*}   {SketchHash, $U=50$} & $\times$ &400  & 0.445  & 0.447 \\ 
    
 & \checkmark    & 9.6 ({41.6}x) & 0.461 & 0.460 (+2\%) \\ 
    
    \hdashline

 \multirow{2}{*}     {SketchHash, $U=100$}&$\times$ &202  & 0.445 &0.446 \\
    
 & \checkmark    & 8.67 (23.2x) & 0.462 &0.460 (+3.1\%) \\
    
	     \hdashline

 \multirow{2}{*}   {SketchHash, $U=1000$} & $\times$ &24  & 0.445  & 0.445 \\ 
    
 & \checkmark    & 8.3 (2.8x) & 0.456 & 0.455 (+2\%) \\ 
    \hline
    \end{tabular}
}
\hfill{}
\vspace{1em}
\caption{
Online hashing results (32 bits) with different update interval parameters ($U$) on the \textbf{LabelMe} dataset. 
All results are averaged from 3 random trials.
For the number of hash table updates, we report the reduction ratio (\eg 8x) for TU.
For  AUC, we report the relative change compared to baseline.
Note: since LabelMe is an unsupervised dataset, the OSH method is not applicable since it requires supervision in the form of class labels.
}
  \label{table:u_labelme}
\end{table*}

%%%%%%%%%%%%%%%%%%%%%%%%%%%%%%%%%%%%%%% PLACES  %%%%%%%%%%%%%%%%%%%%%%%%%%%%%%%%%%%%%%%%%%%%%%%%%%%%%
\begin{table*}[!ht]
%\footnotesize
\small
\centering
{
\hfill{}
\begin{tabular}{l|c|c|c|l}
\hline
\multicolumn{5}{c}{\makecell{\textbf{Places205, 32 bits}}} \\
\hline
\begin{tabular}{@{}c@{}}\textbf{Method} \end{tabular}
& \begin{tabular}{@{}c@{}} \textbf{TU} \end{tabular}
& \begin{tabular}{@{}c@{}} \textbf{HT Updates} \end{tabular}
& \begin{tabular}{@{}c@{}} \textbf{Final mAP} \end{tabular}  
& \begin{tabular}{@{}c@{}}\textbf{AUC (mAP)} \end{tabular} 
\\
\hline
     %\multicolumn{5}{c}{\makecell{OKH}} \\
    \multirow{2}{*}{OKH, $U=50$} & $\times$ & 2001   & 0.182 & 0.163 \\ 
    
    & \checkmark & 8 (250.1x)  & 0.173 & 0.169 (+3.7\%) \\ 
    
    \hdashline

   \multirow{2}{*}{OKH, $U=500$} & $\times$ & 201  & 0.182 & 0.163  \\
    
 & \checkmark    & 7 (28.7x) & 0.165 & 0.161 (-1.2\%)  \\
    
   \hdashline

    \multirow{2}{*}{OKH, $U=5000$} &$\times$ & 21  & 0.182  &  0.156  \\ 
    
 & \checkmark    & 2 (10.5x)  & 0.157 & 0.148 (-5.1\%) \\ 
    \hline
    
    % OSH
    \multirow{2}{*}{OSH, $U=50$} & $\times$ & 2001 & 0.248 & 0.246 \\ 
    
 & \checkmark    & 25 (80x) & 0.239 & 0.238 (-3\%) \\ 
    \hdashline

 \multirow{2}{*}{OSH, $U=500$} & $\times$ & 201  & 0.248 & 0.246  \\
    
 & \checkmark    & 7 (28.7x) & 0.236 & 0.236 (-4.0\%)  \\
    
    \hdashline

 \multirow{2}{*}   {OSH, $U=5000$}&$\times$  & 21 & 0.248 & 0.245  \\ 
    
 & \checkmark    & 2 (10.5x)  & 0.234 &  0.233 (-4\%) \\ 
    
 \hline
      %\multicolumn{5}{c}{\makecell{AdaptHash}} \\
 \multirow{2}{*}   {AdaptHash, $U=50$} & $\times$ & 823.7 & 0.238 & 0.237 \\ 
    
 & \checkmark    & 26.6 (30.8x) & 0.236 & 0.236 (-0.4\%) \\ 
    
    \hdashline

 \multirow{2}{*}  {AdaptHash, $U=500$}  &$\times$  & 200  & 0.238 & 0.237\\
    
 & \checkmark  & 8.6 (23.0x) & 0.236 & 0.236 (-0.4\%) \\
    
    \hdashline

 \multirow{2}{*}   {AdaptHash, $U=5000$} & $\times$ & 21 & 0.238 & 0.237 \\ 
    
 & \checkmark  &  3  (7x) & 0.236 & 0.236 (-0.4\%)\\ 
    
    \hline %\hline
     %\multicolumn{5}{c}{\makecell{SketchHash}} \\
     
 \multirow{2}{*}   {SketchHash, $U=50$} & $\times$ & 2000  & 0.238  & 0.235 \\ 
    
 & \checkmark   & 19.3 (103.4x) & 0.236 & 0.235 (0\%) \\ 
    
    \hdashline

 \multirow{2}{*}  {SketchHash, $U=500$}  &$\times$ & 202  & 0.237 & 0.235  \\
    
 & \checkmark    & 15.3 (13.1x)  & 0.240 & 0.238 ({+1.2\%}) \\
    
    \hdashline

 \multirow{2}{*}   {SketchHash, $U=5000$} &$\times$ & 22  & 0.235 & 0.235  \\ 
    
 & \checkmark   & 6.6 (3.2x) & 0.239  & 0.238 (+1.2\%) \\ 
    
    \hline %\hline
 
    \end{tabular}
}
\hfill{}
\vspace{1em}
\caption{Online hashing results (32 bits) with different update interval parameters ($U$) on the \textbf{Places205} dataset. 
All results are averaged from 3 random trials.
For the number of hash table updates, we report the reduction ratio (\eg 8x) for TU.
For and AUC, we report the relative change compared to baseline.
}
  \label{table:u_places}
\end{table*}

\end{document}